\documentclass[10pt,twocolumn,letterpaper]{article}

\usepackage[pagenumbers]{wacv} %
\usepackage{multirow}
\usepackage{pifont}
\definecolor{lightpurple}{RGB}{200, 160, 255}
\usepackage{tabularx}

\usepackage{algorithm}
\usepackage{algpseudocode}

\usepackage{pgfplots}
\pgfplotsset{compat=1.18}
\usepgfplotslibrary{fillbetween}

\usepackage{colortbl}

\definecolor{wacvblue}{rgb}{0.21,0.49,0.74}
\usepackage[pagebackref,breaklinks,colorlinks,allcolors=wacvblue]{hyperref}

\def\wacvPaperID{650} %
\def\confName{WACV}
\def\confYear{2027}

\title{Closed-Loop Aware Planning via Imitation-Free Speed Scoring for End-to-End Autonomous Driving}

\makeatletter
\newcommand\customfootnotetext[2]{%
  \begingroup
  \renewcommand\thefootnote{#1}%
  \footnotetext{#2}%
  \endgroup
}
\makeatother

\author{Emad Aghajanzadeh*$^{1,2}$, Thomas Schwarzfischer*\textsuperscript{$\dagger$}$^{1,2}$, Marc Kaufeld$^{1}$, Moritz Thoma$^{2}$,\\ Ryan Pégoud\textsuperscript{$\dagger$}$^{2}$, Manoj Rohit Vemparala$^{2}$, Pierpaolo Mori$^{2}$, Nael Fasfous\textsuperscript{$\dagger$}$^{2}$,\\ Johannes Betz$^{1}$, Daniel Mueller-Gritschneder$^{3}$, Ulf Schlichtmann$^{1}$
\\
\normalsize
$^{1}$Technical University of Munich, Germany;
\quad
$^{2}$BMW Group, Germany;
\quad
$^{3}$TU Wien, Austria
\\
{\tt\small \{<firstname>.<lastname>\}$^{1}$@tum.de, $^{2}$@bmw.de, $^{3}$@tuwien.ac.at}
}

\begin{document}
\maketitle

\customfootnotetext{*}{Equal contribution. Co-first authors are ordered in alphabetical order.}
\customfootnotetext{$\dagger$}{Employed at BMW at time of submission.}
\customfootnotetext{}{\copyright\ 2027 IEEE. Personal use of this material is permitted. Permission from IEEE must be obtained for all other uses, in any current or future media.}

\begin{abstract}
End-to-end autonomous driving has predominantly relied on open-loop imitation learning for training, which presents two limitations. First, expert demonstrations show only one behavior per scenario, so the model cannot learn alternative valid behaviors and, in turn, cannot adapt to user-specified preferences at test time. Second, noise or ambiguity in these demonstrations can produce inconsistent, unreliable supervision for the end-to-end models.
Driving is inherently multi-modal: multiple different behaviors can be valid responses to the same scene, and this diversity arises largely from speed rather than path. We introduce a planning paradigm that learns the path through imitation while scoring a set of candidate speeds across multiple criteria with a learned function supervised by physically grounded safety constraints. We realize this paradigm as Closed-Loop Aware Planning for Autonomous Driving (CLAP-AD), a plug-in planning head that integrates with existing end-to-end driving models and, to our knowledge, is the first to enable explicit test-time control over their speed selection without speed-conditioned training. On Bench2Drive-Speed, CLAP-AD sets a new state of the art Speed-Adherence Score of 82.53, topping TFv6, the strongest baseline on this metric, by 12.9 points while retaining strong closed-loop driving performance. On Bench2Drive, it improves driving score by +28.5 and +32.7 over UniAD and VAD, and on NAVSIM~v2, built on real-world driving data, EPDMS by +6.1 over its backbone. Project page: \href{https://clap-ad.github.io}{clap-ad.github.io}.
\end{abstract}
    
\section{Introduction}
\label{sec:intro}

\begin{figure*}[t]
    \centering
    \begin{subfigure}[b]{0.45\textwidth}
        \centering
            \begin{tikzpicture}
        \begin{axis}[
            width=\linewidth,
            height=0.6\linewidth,
            xlabel={Driving Score $\uparrow$},
            ylabel={Speed-Adherence $\uparrow$},
            xmin=10, xmax=100,
            ymin=50, ymax=90,
            grid=major,
            grid style={gray!25, line width=0.3pt},
            axis line style={gray!55, line width=0.5pt},
            tick align=outside, tick pos=left,
            xlabel near ticks, ylabel near ticks,
            tick label style={font=\tiny},
            label style={font=\footnotesize},
            clip=false,
        ]

        \fill[green!55!white, opacity=0.20]
            (axis cs:10,70) rectangle (axis cs:100,90);

        \draw[green!45!black, dashed, semithick]
            (axis cs:10,70) -- (axis cs:100,70);

        \node[anchor=north west, font=\tiny\itshape, text=green!35!black]
            at (axis cs:10,89.3) {controllable};

        \draw[->, thin, gray!55, >=stealth, shorten <=3pt, shorten >=3pt]
            (axis cs:18.90,54.22) -- (axis cs:75.15,79.41);

        \draw[->, thin, gray!55, >=stealth, shorten <=3pt, shorten >=3pt]
            (axis cs:40.23,61.63) -- (axis cs:90.52,75.51);

        \draw[->, thin, gray!55, >=stealth, shorten <=3pt, shorten >=3pt]
            (axis cs:92.59,69.64) -- (axis cs:88.47,82.53);

        \addplot[only marks, mark=*]
            coordinates {(29.66,69.23)}
            node[below,font=\tiny] {TCP-Speed~\cite{shao2026can}};

        \addplot[only marks, mark=*]
            coordinates {(18.90,54.22)}
            node[right,font=\tiny] {UniAD~\cite{hu2023planning}};

        \addplot[only marks, mark=*]
            coordinates {(40.23,61.63)}
            node[right,font=\tiny] {VAD~\cite{jiang2023vad}};

        \addplot[only marks, mark=*]
            coordinates {(92.59,69.64)}
            node[below,font=\tiny] {TFv6~\cite{nguyen2026lead}}; %

        \addplot[only marks, mark=*]
            coordinates {(69.87,68.53)}
            node[below,font=\tiny] {Hydra-NeXt~\cite{li2025hydra}};

        \addplot[only marks, mark=*]
            coordinates {(94.63,51.13)}
            node[above left,font=\tiny] {SparseDriveV2~\cite{sun2026sparsedrivev2}};

        \addplot[only marks, mark=star, mark size=3pt]
            coordinates {(75.15,79.41)}
            node[left,font=\tiny] {CLAP-AD on UniAD};

        \addplot[only marks, mark=star, mark size=3pt]
            coordinates {(90.52,75.51)}
            node[below left,font=\tiny] {CLAP-AD on VAD};

        \addplot[only marks, mark=star, mark size=3pt]
            coordinates {(88.47,82.53)}
            node[above left,font=\tiny] {CLAP-AD on TF};

        \end{axis}
    \end{tikzpicture}
        \caption{Driving Score versus Speed-Adherence.}
        \label{fig:driving_score_speed_adherence}
    \end{subfigure}%
    \hfill
    \begin{subfigure}[b]{0.45\textwidth}
        \centering
        \input{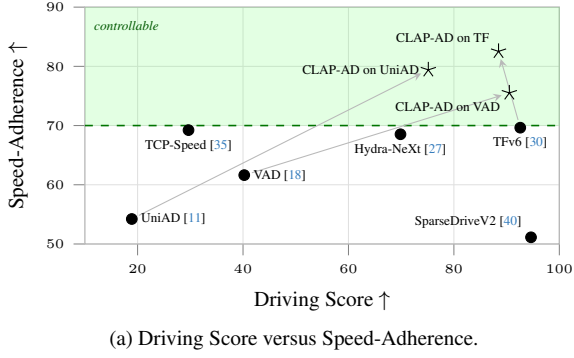}
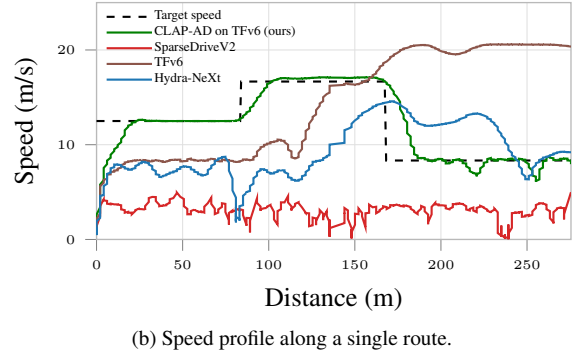
        \caption{Speed profile along a single route.}
        \label{fig:speed_adherence}
    \end{subfigure}
    \caption{\textbf{CLAP-AD improves speed-adherence across three different end-to-end driving models.}
    \subref{fig:driving_score_speed_adherence}~CLAP-AD consistently improves Speed-Adherence across all three backbones; only on TFv6 does this cost a small amount of Driving Score.
    \subref{fig:speed_adherence}~An example of how CLAP-AD respects the user's requested target speed.}
    \label{fig:teaser}
\end{figure*}

Recently, end-to-end (E2E) autonomous driving has gained attention for transforming raw sensor inputs directly into driving decisions within a fully differentiable framework, with approaches such as UniAD~\cite{hu2023planning}, VAD~\cite{jiang2023vad}, and TransFuser~\cite{chitta2022transfuser}. Imitation learning~\cite{le2022survey} is often the preferred paradigm for training such models, favored over reinforcement learning~\cite{sutton1998reinforcement} for enabling stable, efficient open-loop training from expert demonstrations while avoiding complex reward design~\cite{cheng2024pluto, renz2022plant, knox2023reward}. However, driving is inherently a multi-modal task, with multiple valid behaviors possible in the same scenario, and strictly imitating a single expert demonstration prevents the model from learning that alternative behaviors are also valid, contributing to a fundamental mismatch with closed-loop deployment~\cite{zhang2022rethinking, jaeger2025carl}.

To improve closed-loop performance, one line of work disentangles path from speed, observing that a shared temporal-waypoint representation entangles a largely deterministic, lane-constrained path with a genuinely multi-modal speed~\cite{jaeger2023hidden}. Follow-up methods retain this separation but still imitate a single speed value or profile per class~\cite{renz2025simlingo, tang2025hip, wu2026aligndrive}, leaving the multi-modality of driving speed unresolved. A separate line of work scores a fixed trajectory vocabulary against several driving criteria alongside an imitation score~\cite{li2024hydra, li2025hydra, sun2026sparsedrivev2}, yet the evidence for a benefit from these criteria over imitation alone in fully reactive closed-loop settings is mixed at best (\cref{sec:related_work}). Across both lines, path and speed remain optimized under the same objective and architecture, overlooking their fundamentally different nature, and neither offers any support for user test-time controllability over the model's decisions.

Motivated by this, we argue that path and speed should be learned under different objectives, retaining imitation learning for path prediction while replacing expert speed imitation with a multi-criteria scoring mechanism over a set of diverse speed classes. This formulation brings two key advantages. First, evaluating multiple candidate speeds along the predicted path yields a range of safe speed profiles, enabling explicit deployment-time customization of driving speed, a capability largely unexplored in existing E2E driving models. Second, since each criterion is supervised directly from physically grounded constraints rather than a single noisy expert trace, the model learns genuinely safe behavior per criterion, and the resulting scores are directly interpretable, letting us attribute driving failures to particular failures in speed score prediction. To realize this framework, we propose CLAP-AD, a plug-in planning head that integrates into existing E2E driving models: given upstream perception features, it first predicts a driving path, then scores candidate speed classes along it via dedicated, criterion-specific heads. As shown in \cref{fig:teaser}, this design lets CLAP-AD substantially improve how faithfully each backbone follows a user-specified speed while retaining strong closed-loop driving performance.

\begin{enumerate}

    \item We introduce a \textbf{multi-criteria, imitation-free speed learning objective} that explicitly models the suitability of multiple speed behaviors under physically grounded, safety-critical driving criteria, rather than directly imitating a single expert speed. This enables interpretable speed evaluation while better capturing the multi-modal nature of driving speed.
    
    \item We propose \textbf{CLAP-AD}, an extensible plug-in planning head that hierarchically predicts a driving path and uses a \textbf{path-conditioned scoring function} to evaluate candidate speed classes. We integrate CLAP-AD into three backbones: the well-established UniAD and VAD, as well as the recent, strong TFv6. To our knowledge, this is the first approach to bring \textbf{explicit test-time speed control} to existing E2E driving models without speed-conditioned training, allowing users to customize driving speed at deployment while the model continues to respect safety constraints (\cref{fig:speed_adherence}).
    
    \item We conduct extensive closed-loop evaluations on Bench2Drive, the recent Bench2Drive-Speed benchmark and NAVSIM~v2, which is built on real-world driving data. CLAP-AD improves \textbf{Speed-Adherence} on every backbone, by up to $+25.2$ points. Our variants are the only ones in the comparison to reach the highly controllable regime above an adherence of 70 (\cref{fig:driving_score_speed_adherence}). This controllability comes at little to no cost in driving quality: CLAP-AD improves UniAD by +28.5 DS and VAD by +32.7 DS on Bench2Drive and LTFv6 by +6.1 EPDMS on NAVSIM~v2, while on TFv6 it retains driving performance within a few points of the baseline.
\end{enumerate}

\section{Related Work}
\label{sec:related_work}
\subsection{E2E Driving Architectures}
Traditional autonomous driving pipelines decompose the task into independent perception, prediction, and planning modules, limiting the system's ability to recover from upstream perception errors. E2E models address this by jointly training the full pipeline toward a final planning objective, most commonly represented as a temporal trajectory. TransFuser~\cite{chitta2022transfuser} is a widely-used baseline that fuses camera and LiDAR features to regress such a trajectory, while UniAD~\cite{hu2023planning} and VAD~\cite{jiang2023vad} instead adopt query-based and vectorized scene representations, respectively, to better propagate information across modules. Subsequent works refine the architecture: DriveTransformer~\cite{jia2025drivetransformer} lets task queries interact directly with raw sensor features, DiffAD~\cite{wang2025diffad} plans with diffusion over rasterized BEV targets, GenAD~\cite{zheng2024genad} models prediction and planning in a latent trajectory space, DiFSD~\cite{su2024difsd} and SparseDrive~\cite{sun2025sparsedrive} use sparse scene representations, MomAD~\cite{song2025don} adds trajectory and perception momentum, and SpaceDrive~\cite{li2026spacedrive} augments a VLM-based planner with explicit 3D positional encodings. While these architectural advances often improve open-loop metrics, their gains on closed-loop benchmarks such as Bench2Drive~\cite{jia2024bench2drive} remain well below what changes to the representation and training objective subsequently achieved.

\subsection{Planning Representation in E2E Models}
\paragraph{Open-Loop Training Challenge}
Open-loop and closed-loop evaluation are two fundamental paradigms for assessing autonomous driving models~\cite{caesar2020nuscenes, cao2025pseudo, caesar2021nuplan, dosovitskiy2017carla}: open-loop methods predict trajectories from recorded observations without interacting with the environment, while closed-loop methods actively control the vehicle in an interactive simulation. Imitation learning from expert demonstrations on open-loop datasets remains the dominant training paradigm, largely because closed-loop training is computationally expensive, requiring repeated environment rendering and interaction during optimization~\cite{yang2025raw2drive, li2024think2drive}. Yet driving is inherently closed-loop and multi-modal. The vehicle must react to the consequences of its own actions, and multiple valid behaviors can explain the same observation, so a single imitated trajectory learned open-loop often fails to capture the behavior needed for reliable closed-loop control~\cite{jia2024bench2drive, ross2011reduction, li2025hydra}. Despite this, few works have targeted open-loop training strategies tailored for closed-loop performance, relying instead on architectural improvements alone.

\paragraph{Disentangled and Scoring-Based Planning}
As one of the earliest approaches, TF++~\cite{jaeger2023hidden} disentangles path and speed from the commonly used temporal waypoint representation, observing that driving multi-modality mainly arises from speed rather than path; it predicts a separate path and classifies speed over predefined classes, using model uncertainty to interpolate the final value. SimLingo~\cite{renz2025simlingo} is a VLM-based driving model that jointly supports language reasoning and driving, predicting separate temporal and geometric waypoints to alleviate steering ambiguity in joint formulations. Building on this line of work, HiP-AD~\cite{tang2025hip} introduces driving-style waypoints within a hierarchical architecture that jointly predicts temporal, spatial, and driving-style waypoints for finer closed-loop control. In another work, AlignDrive~\cite{wu2026aligndrive} reformulates planning as sequential rather than parallel control prediction, conditioning longitudinal actions on the planned lateral path. While these methods share our disentangled, two-stage formulation, they still imitate a single speed value per class or profile rather than modeling the multi-modality of speed itself, leaving test-time speed controllability unaddressed.

Separately, to incorporate objectives beyond imitation, Hydra-MDP~\cite{li2024hydra} ranks a set of predefined trajectories by their similarity to the expert alongside several driving criteria. Hydra-NeXt~\cite{li2025hydra} further refines this via additional control and refinement branches that sharpen the scoring heads' predictions. Most recently, SparseDriveV2~\cite{sun2026sparsedrivev2} decomposes fixed trajectories into path and speed components and ranks each separately through a coarse-to-fine scoring architecture, scaling the vocabulary to roughly 262K anchors, 32x larger than prior work. While these works explore different scoring architectures, the evidence for a benefit from incorporating driving criteria such as collision in reactive closed-loop settings is mixed at best: Hydra-NeXt's added criteria yield only marginal gains over imitation alone, and SparseDriveV2's strongest results rely on imitation scoring exclusively.

\paragraph{Test-Time Speed Controllability}
One less-explored aspect in E2E models is user customizability over the driving policy, which often appears as driving-speed or driving-style customizability~\cite{shao2026can}. Prior work toward this goal~\cite{li2025learning, pei2026safe, surmann2025multi} operates over privileged agent states and relies on training from scratch or per-style finetuning, assumptions incompatible with E2E models, which plan directly from raw sensor perception under a fixed, imitation-pretrained architecture. To our knowledge, we are the first to provide test-time speed control for existing E2E driving models without speed-conditioned training.

\section{Proposed Method}
\label{sec:proposed_method}
\begin{figure*}[!htpb]
    \centering
    \includegraphics[width=0.95\textwidth]{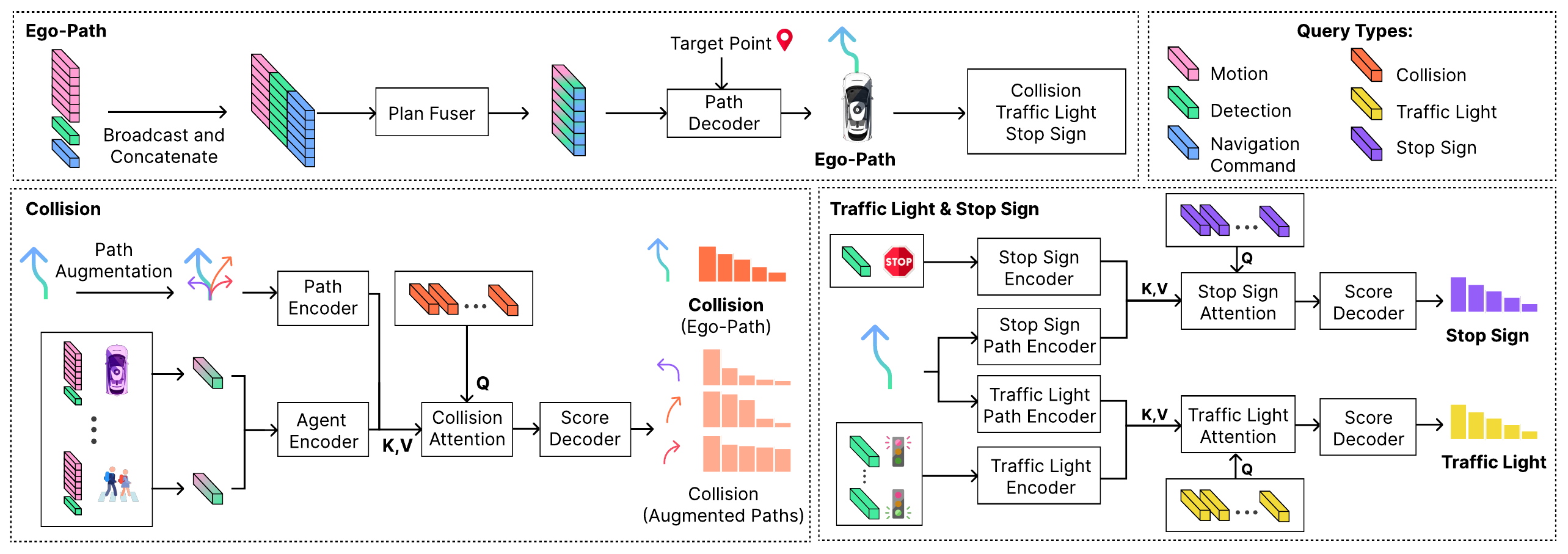}
    \caption{Overview of the CLAP-AD planning head. The architecture consists of an \textit{Ego-Path} block that takes ego detection and motion queries as input and predicts a feasible driving path. The \textit{Collision} block encodes surrounding agents’ detection and motion queries, fuses them with the predicted path representation, and uses learned collision queries to produce collision scores. Similarly, the \textit{Traffic Light} and \textit{Stop Sign} blocks employ dedicated queries that attend to their respective detection features together with the path embedding to generate their speed scores.}
    \label{fig:model_main_figure}
\end{figure*}

While recent works~\cite{zimmerlin2024hidden, renz2025simlingo} have shown that disentangling path and speed is promising, we argue that fully leveraging this decomposition requires distinct learning objectives for each. Driving is inherently multi-modal~\cite{li2024hydra}, and this multi-modality stems largely from speed rather than path~\cite{jaeger2023hidden}: the path is relatively constrained by road structure, while multiple speeds can be equally valid for the same path. We therefore retain imitation learning for path prediction, while introducing an alternative, scoring-based objective for speed to capture its multi-modal nature. We further propose a lightweight planning head that integrates into existing E2E models, enabling closed-loop-aware training in open-loop settings.

\subsection{Training Paradigm}
We formulate speed learning as a multi-modal multi-criteria objective, where a set of predefined speed classes is evaluated against a diverse spectrum of driving requirements. The multi-modality over speed classes allows coverage of diverse urban driving behaviors, while the multi-criteria design explicitly incorporates key safety factors into the learning process. We define a suite of safety-scoring criteria for all speed classes. In this work, we consider three representative criteria: a collision score that reflects whether the ego vehicle keeps sufficient distance to other traffic participants, a traffic light score that reflects whether the required braking for a red light is achievable, and a stop sign score that reflects the same for stop signs.
To optimize the scoring heads, we leverage logged open-loop data to derive training targets for each score; at inference, the network predicts these directly from scene context, without evaluating the underlying rule.

The \textbf{collision} target uses the future positions of surrounding agents from the logged data, so supervision reflects how the scene actually evolved. Given the predicted path, we label each candidate speed with the minimum distance to other actors along that path over a fixed time horizon, excluding static objects, which path prediction already avoids (e.g., a construction obstacle). Since the model's nominal path converges early in training, conditioning collision solely on it limits supervision diversity; we counter this with structured path augmentation, generating neighboring paths and supervising collision on each, which better captures ego–agent interactions under varied motion.

For \textbf{traffic light} and \textbf{stop sign} adherence, we derive each speed's training target from the deceleration required to reach a full stop, $a_{\mathrm{req}}(v) = v^2 / (2\,d_{\mathrm{stop}})$, where $d_{\mathrm{stop}}$ is the distance to the stop line, when one is present ahead. The target decreases smoothly from comfortable braking and reaches 0 once the required deceleration is too high, as in \cref{eq:decel_cases}, so a well-trained head favors speeds that let the ego stop smoothly at the line; with no stop line ahead, all speeds are labeled safe.

\begin{equation}
\label{eq:decel_cases}
    s_{\mathrm{stop}}(v) =
    \begin{cases}
        1, & a_{\mathrm{req}}(v) \le a_{\mathrm{comfort}} \\[6pt]
        \dfrac{a_{\max} - a_{\mathrm{req}}(v)}
              {a_{\max} - a_{\mathrm{comfort}}},
        & a_{\mathrm{comfort}} < a_{\mathrm{req}}(v) < a_{\max} \\[6pt]
        0, & a_{\mathrm{req}}(v) \ge a_{\max} 
\end{cases}
\end{equation}

\subsection{Planning Head Architecture}
We propose a planning head, shown in \cref{fig:model_main_figure}, that operates on a two-stage paradigm: 1) a path planner converting a universal planning query from the upstream model into distance-based path points, and 2) an easily expandable multi-modal multi-criteria speed scoring head that predicts whether predefined speeds are safe given the current scenario. To ensure high adaptability, we designed the path module to handle a range of input modalities. For the example of UniAD, the motion queries are of a multi-modal type (6 motion anchors). We append a navigation command embedding and use an MLP to fuse all input queries into unified plan queries. These are then passed to a GRU together with the target point to auto-regressively predict points along the driving path. The predicted path is passed to the speed head, which consists of three separate branches corresponding to the three scoring criteria. To ensure that the path planner remains focused on spatial navigation and prevent the speed-scoring objectives from influencing the path learning, we detach the predicted path from the computation graph before it enters the speed branches, so speed-scoring gradients cannot propagate back into path prediction. Each branch receives only the information necessary for its criterion to make training more robust and goal-oriented by removing unnecessary scene information. Each branch encodes the predicted path via a dedicated MLP along with score-specific context extracted from upstream detection modules: traffic light and stop sign queries for the respective branches, and all dynamic agent queries (detection and motion) for the collision branch. These inputs are projected into a shared latent space, where learned queries extract relevant task features via cross-attention. Finally, a scoring MLP converts the speed embeddings into logits for each speed class. To accommodate the multi-modal nature of driving, we treat these scores as independent measures of safety rather than a probability distribution.

\subsection{Controllability at Deployment}
\label{sec:controllability_deployment}
A key advantage of our scoring-based design lies in the controllability it affords at deployment. Prior temporal waypoint-based approaches bake a single implicit driving policy into the predicted trajectory, requiring brittle heuristics to recover a desired speed and offering no way to adjust behavior without retraining. In contrast, CLAP-AD outputs explicit, per-criterion safety scores for every speed class, decoupling \textit{what is safe} from \textit{what to do}, so that the desired driving policy can be chosen freely.

We adopt a simple threshold-based strategy in which a speed class is \textit{feasible} if all scoring heads exceed a fixed threshold; how the target speed is then picked among feasible classes determines the driving policy, e.g., the maximum for efficient driving or the median for a defensive style. This same mechanism satisfies a user-specified target speed, as required by Bench2Drive-Speed~\cite{shao2026can}: the selector picks the feasible class closest to the request, re-evaluated at every step, so the vehicle tracks the request whenever safe and falls back automatically when the scenario demands it, e.g., a lead vehicle or red light. If no class is feasible, the selector falls back to $0\,m/s$, i.e., the vehicle brakes. Notably, this requires no architectural changes or retraining: the network always predicts the same safety scores, and the desired speed only biases selection at inference time.

Finally, we convert the predicted path and the selected target speed into low-level driving commands via two independent controllers: a lateral PID controller that derives the steering angle from the predicted path, and a longitudinal PID controller that derives the throttle and brake from the selected target speed and the current ego speed.

\section{Experiments}
\label{sec:experiments}

\definecolor{pdmColor}{rgb}{1.0000,0.4980,0.0549}
\definecolor{t2dColor}{rgb}{0.5804,0.4039,0.7412}

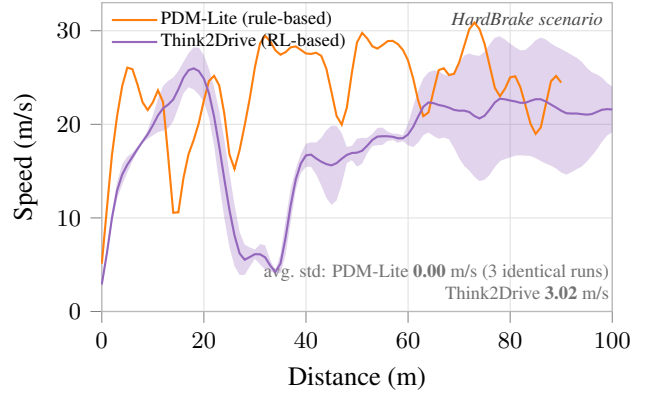
\begin{figure}[t]
    \centering
    \begin{tikzpicture}
    \begin{axis}[
        xlabel={Distance (m)},
        ylabel={Speed (m/s)},
        tick label style={font=\small},
        label style={font=\small},
        xmin=0, xmax=100,
        ymin=0, ymax=33,
        width=\linewidth, height=0.68\linewidth,
        grid=major, grid style={gray!25, line width=0.3pt},
        axis line style={gray!55, line width=0.5pt},
        tick align=outside, tick pos=left,
        xlabel near ticks, ylabel near ticks,
        legend style={
            at={(0,1)},
            anchor=north west,
            font=\scriptsize,
            nodes={inner sep=1pt},
            cells={anchor=west},
            row sep=0pt,
            column sep=0pt,
            draw=none,
            fill=none,
        },
        legend cell align={left},
        thick,
    ]

    \node[anchor=north east, font=\scriptsize\itshape, text=gray!55!black]
        at (axis cs:100,33) {HardBrake scenario};

    \node[anchor=south east, font=\scriptsize, align=right, text=black!55, inner sep=1pt]
        at (axis cs:100,1) {avg.\ std: PDM-Lite \textbf{0.00} m/s (3 identical runs)\\Think2Drive \textbf{3.02} m/s};

    \addplot[name path=pdmUpper, draw=none, forget plot]
        table[x=distance, y=upper]{images/expert_speed_plot/tikz_data/pdm_lite_mean_std.dat};
    \addplot[name path=pdmLower, draw=none, forget plot]
        table[x=distance, y=lower]{images/expert_speed_plot/tikz_data/pdm_lite_mean_std.dat};
    \addplot[draw=none, fill=pdmColor, fill opacity=0.28, forget plot]
        fill between[of=pdmUpper and pdmLower];
    \addplot[color=pdmColor, thick, solid, mark=none]
        table[x=distance, y=mean]{images/expert_speed_plot/tikz_data/pdm_lite_mean_std.dat};
    \addlegendentry{PDM-Lite (rule-based)}

    \addplot[name path=t2dUpper, draw=none, forget plot]
        table[x=distance, y=upper]{images/expert_speed_plot/tikz_data/think2drive_mean_std.dat};
    \addplot[name path=t2dLower, draw=none, forget plot]
        table[x=distance, y=lower]{images/expert_speed_plot/tikz_data/think2drive_mean_std.dat};
    \addplot[draw=none, fill=t2dColor, fill opacity=0.28, forget plot]
        fill between[of=t2dUpper and t2dLower];
    \addplot[color=t2dColor, thick, solid, mark=none]
        table[x=distance, y=mean]{images/expert_speed_plot/tikz_data/think2drive_mean_std.dat};
    \addlegendentry{Think2Drive (RL-based)}

    \end{axis}
    \end{tikzpicture}
    \caption{\textbf{Speed-profile variability across two experts on a repeated HardBrake scenario.} Lines show the mean speed over three re-runs of the same route; shaded bands show $\pm 1$ standard deviation. PDM-Lite and Think2Drive brake at markedly different points along the route, showing that even expert-level speed behavior is inherently multi-modal. Within each expert, PDM-Lite (rule-based) reproduces an identical profile across all three runs (avg.\ std $=0.00$~m/s), while Think2Drive (RL-based) diverges substantially between runs (avg.\ std $=3.02$~m/s).}
    \label{fig:expert_speed_noise}
\end{figure}

\subsection{Benchmarks and Metrics}
\label{sec:benchmarks}
We evaluate on three benchmarks. \textbf{Bench2Drive}~\cite{jia2024bench2drive} provides 1,000 open-loop clips of expert driving~\cite{li2024think2drive} across 44 CARLA~\cite{dosovitskiy2017carla} scenarios for training, and 220 closed-loop test routes covering challenging edge cases; Dev10~\cite{jia2025drivetransformer} is a 10-route subset used for fast iteration. We follow the official protocol and report Driving Score (DS), Success Rate (SR) and Efficiency. We do not report Comfort, as measurement issues with this metric have been identified in~\cite{aghajanzadeh2026comfortablystuckevaluationpitfalls} and are summarized in the supplementary. \textbf{Bench2Drive-Speed}~\cite{shao2026can} extends this setting by supplying a user-specified desired speed at every step. Over its 48 routes, split into Easy, Medium and Hard by how demanding the request is, the \textbf{Speed-Adherence (SA)} score measures how closely the realized ego speed tracks the request, while the standard metrics continue to penalize infractions; the two together characterize the trade-off between controllability and safe driving. For baselines that accept no speed request, SA measures how often their own fixed driving speed happens to coincide with the request, so it serves as a reference level rather than a measure of controllability. Finally, \textbf{NAVSIM~v2}~\cite{cao2025pseudo} provides a real-world pseudo-simulation evaluation, summarized by the Extended Predictive Driver Model Score (EPDMS), which aggregates hard safety constraints with soft driving-quality objectives; we define its sub-metrics in the supplementary.

\definecolor{gainGreen}{RGB}{0,120,60}
\definecolor{dropRed}{RGB}{178,45,45}
\providecommand{\gain}[1]{{\scriptsize\textcolor{gainGreen}{$+$#1}}}
\providecommand{\drop}[1]{{\scriptsize\textcolor{dropRed}{$-$#1}}}

\definecolor{saBand}{rgb}{0.89,1.00,0.89}
\providecommand{\sahi}[1]{\cellcolor{saBand}#1}

\begin{table*}[t]
    \centering
    \caption{\textbf{Bench2Drive-Speed results on 48 routes.} Speed-Adherence (SA) measures how faithfully a policy follows the user-specified speed. Each backbone sits directly above its CLAP-AD variant; inline values give the change relative to it. Best per column in \textbf{bold}. \colorbox{saBand}{Shaded} cells mark Speed-Adherence above 70, the controllable range highlighted in \cref{fig:driving_score_speed_adherence}.}
    \label{tab:b2d_speed}
    \small
    \setlength{\tabcolsep}{8pt}
    \renewcommand{\arraystretch}{1.15}
    \begin{tabular}{l cccc ccc}
        \toprule
        & \multicolumn{4}{c}{\textbf{Speed-Adherence $\uparrow$}}
        & \multicolumn{3}{c}{\textbf{Traditional Metrics $\uparrow$}} \\
        \cmidrule(lr){2-5} \cmidrule(lr){6-8}
        \textbf{Method}
        & \textbf{All} & \textbf{Easy} & \textbf{Medium} & \textbf{Hard}
        & \textbf{Efficiency} & \textbf{Driving Score} & \textbf{Success Rate} \\
        \midrule
        TCP-Speed~\cite{shao2026can} & 69.23 & \sahi{76.18} & 68.71 & 62.81 & 169.84 & 29.66 & 18.75 \\
        Hydra-NeXt~\cite{li2025hydra} & 68.53 & 68.68 & 67.94 & 68.96 & 157.05 & 69.87 & 39.58 \\
        SparseDriveV2~\cite{sun2026sparsedrivev2} & 51.13 & 52.50 & 49.72 & 51.18 & 91.77 & \textbf{94.63} & 60.41 \\
        \midrule
        UniAD~\cite{hu2023planning}     & 54.22 & 56.14 & 52.62 & 53.90 & 108.49 & 18.90 & 4.16 \\
        \quad + CLAP-AD (ours) & \sahi{79.41~\gain{25.2}} & \sahi{83.83} & \sahi{75.40} & \sahi{79.00} & 170.03 & 75.15~\gain{56.3} & 41.67 \\
        \cmidrule(lr){1-8}
        VAD~\cite{jiang2023vad}       & 61.63 & 64.64 & 62.60 & 57.66 & 130.53 & 40.23 & 20.83 \\
        \quad + CLAP-AD (ours)   & \sahi{75.51~\gain{13.9}} & \sahi{80.42} & \sahi{72.05} & \sahi{74.07} & 159.48 & 90.52~\gain{50.3} & 52.08 \\
        \cmidrule(lr){1-8}
        TFv6~\cite{nguyen2026lead}      & 69.64 & \sahi{70.75} & 68.30 & 69.86 & 154.58 & 92.59 & 58.33 \\
        \quad + CLAP-AD (ours)  & \sahi{\textbf{82.53}~\gain{12.9}} & \sahi{\textbf{89.79}} & \sahi{\textbf{76.81}} & \sahi{\textbf{80.98}} & \textbf{187.80} & 88.47~\drop{4.1} & 50.00 \\
        \bottomrule
    \end{tabular}
\end{table*}

\definecolor{gainGreen}{RGB}{0,120,60}
\definecolor{dropRed}{RGB}{178,45,45}

\definecolor{saBand}{rgb}{0.89,1.00,0.89}

\providecommand{\cmark}{\textcolor{gainGreen}{\ding{51}}}
\providecommand{\xmark}{\textcolor{black!35}{\ding{55}}}
\providecommand{\gain}[1]{{\scriptsize\textcolor{gainGreen}{$+$#1}}}
\providecommand{\drop}[1]{{\scriptsize\textcolor{dropRed}{$-$#1}}}

\begin{table}[t]
\centering
\small
\setlength{\tabcolsep}{4pt}
\caption{\textbf{Closed-loop results on the 220 Bench2Drive routes.} Methods are grouped by training expert, each backbone directly above its CLAP-AD variant. \textbf{Ctrl.}: supports user-specified speed at test time. $\Delta$DS is relative to the method's own backbone. $\dagger$ from the re-implementation of~\cite{song2025don}. \colorbox{saBand}{Shaded} rows are the only methods offering test-time speed control; \cref{tab:b2d_speed} reports the Speed-Adherence gained in exchange for the small DS cost on TFv6.}
\label{tab:main_results_bench2drive}
\resizebox{\columnwidth}{!}{%
\begin{tabular}{lcccc}
\toprule
\textbf{Method}
& \textbf{Ctrl.}
& \textbf{DS $\uparrow$}
& \textbf{$\Delta$DS}
& \textbf{SR (\%) $\uparrow$} \\
\midrule
\multicolumn{5}{l}{\textit{Bench2Drive-base data (Think2Drive~\cite{li2024think2drive} expert)}} \\
SparseDrive$\dagger$~\cite{sun2025sparsedrive}    & \xmark & 44.54 & -- & 16.71 \\
GenAD~\cite{zheng2024genad}                       & \xmark & 44.81 & -- & 15.90 \\
MomAD~\cite{song2025don}                          & \xmark & 47.91 & -- & 18.11 \\
DiFSD~\cite{su2024difsd}                          & \xmark & 52.02 & -- & 21.00 \\
Hydra-MDP~\cite{li2024hydra}                      & \xmark & 59.95 & -- & 29.82 \\
DiffAD~\cite{wang2025diffad}                      & \xmark & 67.92 & -- & 38.64 \\
Hydra-NeXt~\cite{li2025hydra}                     & \xmark & 73.86 & -- & 50.00 \\
HiP-AD~\cite{tang2025hip}                         & \xmark & 86.77 & -- & 69.09 \\
AlignDrive~\cite{wu2026aligndrive}                & \xmark & 89.07 & -- & 73.18 \\
SparseDriveV2~\cite{sun2026sparsedrivev2}         & \xmark & 89.15 & -- & 70.00 \\
\cmidrule(lr){1-5}
UniAD~\cite{hu2023planning}                       & \xmark & 45.81 & -- & 16.36 \\
\rowcolor{saBand} \quad + CLAP-AD (ours)                            & \cmark & 74.29 & \gain{28.5} & 49.09 \\
VAD~\cite{jiang2023vad}                           & \xmark & 42.35 & -- & 15.00 \\
\rowcolor{saBand} \quad + CLAP-AD (ours)                            & \cmark & 75.04 & \gain{32.7} & 47.27 \\
\midrule
\multicolumn{5}{l}{\textit{Rule-based expert (PDM-Lite~\cite{sima2024drivelm, beisswenger2024pdm})}} \\
SpaceDrive~\cite{li2026spacedrive}                  & \xmark & 78.02 & -- & 55.11 \\
SimLingo~\cite{renz2025simlingo}                  & \xmark & 85.07 & -- & 67.27 \\
TFv6~\cite{nguyen2026lead}                        & \xmark & \textbf{94.70} & -- & \textbf{82.1} \\
\rowcolor{saBand} \quad + CLAP-AD (ours)                            & \cmark & 92.30 & \drop{2.4} & 75.00 \\
\bottomrule
\end{tabular}%
}
\end{table}

\subsection{Training Setup}
We integrate CLAP-AD into three structurally different frameworks: UniAD, which adopts a fully query-based transformer design; VAD, which relies on a vectorized scene representation while still maintaining object- and motion-level tokens; and TFv6~\cite{nguyen2026lead}, the most recent of the three, a multi-modal camera--LiDAR architecture that exposes a dense BEV feature map rather than the object- and motion-level tokens UniAD and VAD provide. In all cases we replace the original planning head with CLAP-AD, which transfers across both sparse query-based and dense BEV scene encodings without major architectural changes. Similar to related work~\cite{nguyen2026lead, hu2023planning, jiang2023vad}, the perception modules are pre-trained before CLAP-AD is attached, so the detection and motion queries feeding our scoring branches are already reliable when E2E training begins, and each backbone otherwise follows the schedule established for it in prior work. TFv6 additionally consumes LiDAR alongside camera input, which enriches the geometric evidence available to our collision branch. The rest of design choices and training hyperparameters are provided in the supplementary. 

Our experiments span three sources of expert supervision that differ not only in quality but in kind. UniAD and VAD are trained on Bench2Drive-base, whose demonstrations come from Think2Drive~\cite{li2024think2drive}, a privileged expert that embeds a world model and is trained via \emph{reinforcement learning}. TFv6 is trained on LEAD~\cite{nguyen2026lead}, which uses a privileged \emph{rule-based} PDM-Lite expert~\cite{sima2024drivelm, beisswenger2024pdm} restricted to what the student's own cameras can see. Think2Drive's learning-based nature introduces prediction noise during data collection that is largely absent from LEAD; we quantify this on a repeated scenario in \cref{fig:expert_speed_noise}. Our NAVSIM experiments (\cref{sec:navsim}) further train an LTFv6 backbone on NAVSIM's human driving demonstrations, whose speed choices carry the everyday variability of real drivers. Evaluating across all three lets us assess CLAP-AD under supervision of markedly different consistency.

\subsection{Speed Adherence on Bench2Drive-Speed}
\label{sec:speed_adherence}

Bench2Drive-Speed is the only benchmark that measures both aspects of what CLAP-AD provides, reporting closed-loop Driving Score alongside how faithfully a policy follows a user-specified speed. As described in \cref{sec:controllability_deployment}, serving a requested speed $v_{\mathrm{req}}$ requires no retraining: the controller simply picks the safe class closest to it. We now evaluate how well this holds up in practice.

\cref{tab:b2d_speed} reports the results. CLAP-AD improves Speed-Adherence on every backbone (+25.2 on UniAD, +13.9 on VAD and +12.9 on TFv6) and keeps this advantage across all three difficulty splits. The Hard split is the most telling, as it demands speed adherence amid the most complicated traffic situations. A planner that is only superficially controllable would fall back on a safe default here and lose adherence, whereas CLAP-AD raises TFv6 from 69.86 to 80.98. On the two Bench2Drive-trained backbones the gain in adherence arrives together with a large gain in driving quality, raising UniAD from 18.90 to 75.15 DS and VAD from 40.23 to 90.52 DS.

This trade-off is clearest against TFv6 and SparseDriveV2, the two strongest baselines in \cref{tab:b2d_speed}. CLAP-AD attains the best Speed-Adherence of any evaluated method (82.53), trading only 4.1 DS against its own TFv6 baseline (88.47 versus 92.59) for that gain, an operating point we regard as preferable for a deployed system: the baseline's higher score comes from a policy with no mechanism to accept a speed request at all, so its 69.64 SA simply reflects whatever speed its imitation objective happened to learn. The comparison against SparseDriveV2 is an even sharper version of the same trade: CLAP-AD gives up 6.2 DS (88.47 versus 94.63) for a 31.4-point jump in Speed-Adherence (82.53 versus 51.13, the lowest of any evaluated method), since SparseDriveV2 reaches its driving score through a generally cautious policy (the lowest Efficiency in the table, 91.77) rather than by satisfying the request. CLAP-AD also exceeds TCP-Speed~\cite{shao2026can} by 13.3 SA points at nearly three times its Driving Score, even though TCP-Speed is purpose-built for desired-speed conditioning and trained on a dedicated speed-conditioned dataset.

\subsection{Closed-Loop Driving Performance on Bench2Drive}
\label{sec:b2d_results}
\cref{tab:main_results_bench2drive} compares CLAP-AD across three backbones against their respective baselines, under two training regimes that differ in the expert producing the demonstrations. These 220 routes differ in type and scenarios from the 48-route Bench2Drive-Speed suite used in \cref{sec:speed_adherence}, so absolute driving scores should be compared within each table rather than across the two.
On the Bench2Drive-base training data, whose demonstrations come from the reinforcement-learned Think2Drive expert~\cite{li2024think2drive}, CLAP-AD improves UniAD by +28.5 DS and VAD by +32.7 DS, elevating both to a competitive level (74--75 DS) among much more recent approaches and without extended training schedules or architectural scaling. Both variants also surpass methods that redesign the planning architecture itself, including SparseDrive~\cite{sun2025sparsedrive} (44.54 DS), Hydra-MDP~\cite{li2024hydra} (59.95 DS) and Hydra-NeXt~\cite{li2025hydra} (73.86 DS), even though we leave the backbone untouched and change only the planning objective and head.

On TFv6, CLAP-AD reaches 92.30 DS and 75.00 SR, 2.4 points below its own TFv6 baseline (94.70 DS), the only backbone where CLAP-AD does not improve on the baseline it is attached to. It nonetheless surpasses every other Bench2Drive-trained method in \cref{tab:main_results_bench2drive}, including AlignDrive (89.07 DS) and SparseDriveV2 (89.15 DS).
This gap traces back to how ambiguous each expert's speed labels are, which is what our scoring objective is designed to address. PDM-Lite is a privileged, rule-based planner that computes speed deterministically from ground-truth scene state, so similar scenes receive consistent labels and imitation already provides a strong training signal. Think2Drive, by contrast, is reinforcement-learned and produces inconsistent speeds across visually similar scenes, which is the kind of noise our objective is designed to handle, as reflected in the large UniAD and VAD gains above. Since privileged rule-based experts exist only in simulation, real-world deployment is always closer to the noisier, Think2Drive-like regime, making this the setting that matters most in practice. Qualitative examples of the per-criterion scores are provided in the supplementary.

\subsection{Evaluation on NAVSIM~v2}
\label{sec:navsim}

\begin{table}[t]
\centering
\caption{Results on the \textbf{NAVSIM~v2 navhard} benchmark~\cite{cao2025pseudo}. CLAP-AD replaces only the planning head of \textbf{LTFv6} and shares its backbone, so the \colorbox{saBand}{shaded} rows should be read directly against the LTFv6 rows above. As in \cref{tab:main_results_bench2drive}, shading marks the model offering test-time speed control. Results for LEAD-LTFv6 are taken from the public NAVSIM~v2 leaderboard.}
\label{tab:navsim_results}
\definecolor{saBand}{rgb}{0.89,1.00,0.89}
\renewcommand\arraystretch{0.9}
\tabcolsep=1.0mm
\resizebox{\columnwidth}{!}{
\begin{tabular}{l c ccccc ccccc}
\toprule
Method & Stage & NC$\uparrow$& DAC$\uparrow$& DDC$\uparrow$& TL$\uparrow$& EP$\uparrow$& TTC$\uparrow$& LK$\uparrow$& HC$\uparrow$& EC$\uparrow$& EPDMS$\uparrow$ \\
\midrule

\multirow{2}{*}{TransFuser~\cite{chitta2022transfuser}}
& Stage 1 & 96.2 & 79.5 & 99.1 & 99.5 & 84.1 & 95.1 & 94.2 & 97.5 & \textbf{79.1} & \multirow{2}{*}{23.1} \\
& Stage 2 & 77.7 & 70.2 & 84.2 & 98.0 & 85.1 & 75.6 & 45.4 & 95.7 & \textbf{75.9} &  \\
\addlinespace%

\multirow{2}{*}{DiffusionDrive~\cite{liao2025diffusiondrive}}
& Stage 1 & 96.0 & 79.7 & 97.4 & 99.5 & 81.3 & 93.1 & 90.8 & 96.8 & 73.8 & \multirow{2}{*}{24.2} \\
& Stage 2 & 82.1 & 72.2 & 88.5 & 98.7 & 85.1 & 78.8 & 49.2 & 89.3 & 71.2 &  \\
\addlinespace%

\multirow{2}{*}{GuideFlow~\cite{liu2026guideflow}}
& Stage 1 & 96.6 & 80.5 & 96.3 & 99.3 & 82.3 & 94.9 & 91.5 & 97.7 & 67.8 & \multirow{2}{*}{27.1} \\
& Stage 2 & 87.3 & 76.7 & 88.8 & 99.2 & 84.3 & 85.1 & 49.7 & 93.1 & 44.5 &  \\
\addlinespace%

\multirow{2}{*}{Senna-E2E~\cite{jiang2024senna}}
& Stage 1 & 95.6 & 86.0 & 98.9 & 99.6 & 83.9 & 95.1 & \textbf{95.3} & 97.6 & 75.6 & \multirow{2}{*}{27.2} \\
& Stage 2 & 78.6 & 74.8 & 84.8 & 98.2 & 88.2 & 75.7 & 46.9 & 96.0 & 65.8 &  \\
\addlinespace%

\multirow{2}{*}{LTFv6~\cite{nguyen2026lead}}
& Stage 1 & 96.6 & 86.7 & \textbf{99.2} & \textbf{99.6} & \textbf{84.5} & 95.1 & 94.4 & 97.8 & 76.4 & \multirow{2}{*}{31.9} \\
& Stage 2 & 79.9 & 75.6 & 86.3 & 97.9 & \textbf{89.6} & 76.1 & 50.1 & 95.2 & 66.7 &  \\
\addlinespace%

\rowcolor{saBand}
& Stage 1 & \textbf{98.8} & \textbf{87.6} & 99.0 & \textbf{99.6} & 78.2 & \textbf{98.0} & 93.1 & \textbf{98.0} & 37.8 & \\
\rowcolor{saBand}
\multirow{-2}{*}{\textbf{LTFv6} + CLAP-AD (ours)}
& Stage 2 & \textbf{90.3} & \textbf{79.1} & \textbf{89.6} & \textbf{99.2} & 78.8 & \textbf{87.7} & \textbf{53.6} & \textbf{97.2} & 51.6 & \multirow{-2}{*}{\textbf{38.0}} \\

\bottomrule

\end{tabular}
}
\end{table}

To study the applicability of CLAP-AD beyond simulation, we report results on NAVSIM~v2 using a Latent Transfuser (LTFv6) backbone, replacing its planning head with CLAP-AD and restricting scoring to the collision head. Because NAVSIM evaluates a single four-second trajectory without intermediate replanning, we extend the planning horizon to $48 m$ and the collision horizon to the full $4 s$, and replace the hard speed thresholding used in CARLA with a confidence-weighted average over speed bins, which is then integrated into a temporal trajectory along the predicted path. The supplementary gives the full procedure.

As shown in \cref{tab:navsim_results}, CLAP-AD improves its LTFv6~\cite{nguyen2026lead} backbone by +6.1 EPDMS (31.9 to 38.0), the highest score among the compared methods. The gains land precisely on the criterion our collision head optimizes: no-at-fault collision rises from 96.6 to 98.8 in stage~1 and from 79.9 to 90.3 under the reactive stage~2 simulation, and time-to-collision from 95.1 to 98.0 and from 76.1 to 87.7, both the best in the table by a clear margin ($+3.0$ NC and $+2.6$ TTC over the next-best method in stage~2), with drivable-area compliance improving alongside them (75.6 to 79.1). This comes at a cost in ego progress (84.5 to 78.2 in stage~1, 89.6 to 78.8 in stage~2) and extended comfort (76.4 to 37.8 and 66.7 to 51.6, respectively). That the same distance-based collision criterion we defined in CARLA transfers to real sensor data without redefinition supports our claim that it encodes a physically grounded driving constraint rather than a benchmark-specific heuristic. NAVSIM demonstrations come from human driving rather than a privileged planner, so its speed supervision carries the same ambiguity as the Think2Drive regime, and the result confirms our hypothesis: when the expert demonstrations are noisy, whether from a reinforcement-learning expert such as Think2Drive or from human drivers, CLAP-AD helps the model learn more robust driving behavior, whereas against a rule-based expert such as PDM-Lite it remains comparable to direct imitation. Real-world data always contains such ambiguity. The comparison is at a small-to-medium backbone scale; approaches with substantially larger perception backbones, such as DriveFuture~\cite{hong2026drivefuture} and SimScale~\cite{tian2026simscale} with V2-99~\cite{lee2020centermask}, report higher EPDMS, so we position CLAP-AD as highly competitive within this capacity regime rather than as state of the art.

\subsection{Ablation: Where the Gains Come From}
\label{sec:ablation}

\begin{table}[t]
\centering
\small
\setlength{\tabcolsep}{4pt}
\caption{\textbf{Ablation on UniAD, evaluated on Dev10.} The upper block varies only the speed objective inside UniAD's original planning head; the lower block keeps our scoring objective and replaces the head with CLAP-AD. Scoring only pays off once the head is designed to consume it (40.50 $\rightarrow$ 59.44).}
\label{tab:uniad_incremental}
\resizebox{\columnwidth}{!}{%
\begin{tabular}{llcc}
\toprule
\textbf{Speed objective} & \textbf{Planning head} & \textbf{DS $\uparrow$} & \textbf{SR (\%) $\uparrow$} \\
\midrule
Joint temporal wp. (baseline)          & UniAD   & 45.12 & 20 \\
Path + speed classification~\cite{jaeger2023hidden}     & UniAD   & 43.03 & 0  \\
Dual spatial--temporal wp.~\cite{renz2025simlingo, tang2025hip}  & UniAD   & 56.44 & 20 \\
Multi-criteria scoring (ours)          & UniAD   & 40.50 & 0  \\
\midrule
Multi-criteria scoring (ours)          & CLAP-AD & 59.44 & 30 \\
\quad + path aug. \& horizon (12\,m / 1\,s) & CLAP-AD & 72.71 & 50 \\
\quad + per-head path encoding         & CLAP-AD & \textbf{75.19} & \textbf{50} \\
\quad + longer horizon (24\,m / 2\,s)  & CLAP-AD & 72.30 & 40 \\
\bottomrule
\end{tabular}%
}
\end{table}

We now isolate the two ingredients behind these results (the speed objective and the planning head) using UniAD on Dev10 (\cref{tab:uniad_incremental}).

\paragraph{The speed objective alone is not enough.}
The upper block of \cref{tab:uniad_incremental} keeps UniAD's original planning head and varies only how speed is learned. Replacing joint temporal waypoints with a path plus a speed classifier~\cite{jaeger2023hidden} does not help (45.12 $\rightarrow$ 43.03), whereas predicting dual spatial and temporal waypoints~\cite{renz2025simlingo, tang2025hip} raises DS to 56.44, confirming that separating path from speed removes ambiguity. Substituting our multi-criteria scoring objective into the same head, however, \emph{drops} DS to 40.50: UniAD's planning head collapses all scene context into a single trajectory query and cannot route the criterion-specific evidence that scoring requires.

\paragraph{The head unlocks it.}
Keeping the identical objective and replacing only the head with CLAP-AD, which allocates dedicated queries per scoring criterion, lifts DS from 40.50 to 59.44, above every alternative speed representation and $+14$ over the UniAD baseline. This is the central result of the ablation: CLAP-AD realizes the scoring paradigm effectively by giving each criterion its own query and evidence path. We further experiment three design choices on it. Augmenting each path with four structured perturbations (two constant lateral shifts of $\pm 0.6 m$ and two progressive drifts growing to $\pm 0.8 m$) exposes the collision head to plausible cut-in deviations rather than only the nominal path, and extending the path horizon from $6 m$ to $12 m$ and the collision horizon from $0.5 s$ to $1.0 s$ leaves enough reaction time for junction entry, overtaking, and emergency braking; together these add $+13$ DS. Per-head path encoders, since each criterion reads the path differently, add a further $+2.5$ DS to reach 75.19. Pushing the horizons to $24 m$ and $2 s$ does not help (72.30): anticipating agents two seconds ahead from a single observation makes the supervision noisier and less discriminative between neighboring speed classes.

\section{Conclusion}
\label{sec:conclusion}
In this paper, by building upon the success of disentangled path and speed representations, we propose to train them under different objectives. Motivated by the inherently multi-modal nature of driving speed, we introduce an imitation-free learning objective that moves beyond strict expert imitation, and design CLAP-AD, a lightweight planning head that implements a query-based, multi-criteria scoring mechanism to realize it.

CLAP-AD is, to our knowledge, the first approach to bring explicit test-time speed control to existing E2E driving models without speed-conditioned training, improving Speed-Adherence over TFv6, the strongest baseline on this metric, by 12.9 points on Bench2Drive-Speed. It also improves closed-loop driving on UniAD and VAD by +28.5 and +32.7 Driving Score, stays within a few points of TFv6's strong baseline on Bench2Drive, and surpasses its LTFv6 backbone by +6.1 EPDMS on NAVSIM~v2, which is built on real-world driving data, suggesting that it applies beyond simulation.

Ablations highlight the advantages of scoring-based speed modeling, notably its interpretability and deployment flexibility: explicit, per-criterion safety scores simplify controller design and make the model's decisions easier to audit, improving the practical deployability of E2E driving systems. Limitations are discussed in the supplementary.
{
    \small
    \bibliographystyle{ieeenat_fullname}
    \bibliography{main}
}

\clearpage
\appendix
\ifdefined\suppstandalone\else\section*{Supplementary Material}\fi
This supplementary material is organized as follows. \cref{sec:supp_architecture} specifies the architecture and hyperparameters of the planning head, together with its size on TFv6 and the two adaptations we make to the perception backbone so that it supplies the queries our scoring branches consume. We then report additional results: \cref{sec:controllability} analyzes how the controller reads the learned scores at test time and the range of driving styles this exposes, and \cref{sec:supp_ability} gives a per-ability breakdown on Bench2Drive. \cref{sec:supp_collision} defines how the collision ground truth is computed for each speed class, and \cref{sec:supp_qualitative} illustrates the resulting per-criterion scores on individual CARLA scenes. Finally, we describe the evaluation protocols: \cref{sec:supp_metrics} defines the Bench2Drive closed-loop metrics, \cref{sec:comfort} explains why we omit the Comfort metric from the comparisons in the main paper, and \cref{sec:navsim_supp} covers the NAVSIM~v2 sub-metrics together with the procedure that converts CLAP-AD's outputs into a NAVSIM-compatible trajectory. \cref{sec:limitations} closes with the limitations of the scoring approach.

\section{Model Architecture}
\label{sec:supp_architecture}
\subsection{Planning Head}

\paragraph{Branch architecture.} The detailed architecture of CLAP-AD is summarized in \cref{tab:clap_architecture_detailed}. All modules operate on a feature embedding of dimension 256. The ego–path block first fuses scene features using a lightweight MLP and predicts the driving path through a GRU-based decoder conditioned on the target point. The collision branch encodes position and motion of surrounding agents as well as the predicted path to assess potential collision risks, using an attention module followed by a score decoder that estimates a collision safety score for each speed class. The traffic light and stop sign branches process static traffic control elements using a similar encoder–attention–decoder structure to evaluate speed suitability with respect to traffic rules.

\paragraph{Per-backbone adaptations.} For both UniAD and VAD, the described architecture applies, with the only difference being the fusion dimension (\texttt{fuser\_dim}): 3 for UniAD and 4 for VAD. This difference arises from the shape of their motion queries: UniAD assigns the same embedding size to motion queries as used for detection queries, whereas VAD allocates twice the embedding size for its motion queries, resulting in a larger input feature dimension. TFv6 needs a different treatment altogether, because its perception is not query-based. Instead of supplying object- and motion-level tokens, it produces a dense BEV feature map, which we pass in whole and pool into the fixed set of tokens each CLAP-AD branch expects. Following the design of its original planning head, we also keep that head's linear layer in place of the GRU decoder and condition the path on three target points rather than one, so that CLAP-AD stays compatible with the interface the baseline was built around.

\begin{table}[t]
\centering
\footnotesize
\setlength{\tabcolsep}{3pt}
\renewcommand{\arraystretch}{0.95}
\begin{tabular}{p{4.2cm} p{4.3cm}}
\toprule
\textbf{Module} & \textbf{Architecture} \\
\midrule

\multicolumn{2}{l}{\textbf{Ego--Path Block}} \\
\midrule

Plan Fuser & FC($256\!\cdot\!fuser\_dim$, 256) \\
Path Decoder & TP\_FC(2,256), GRU(256,256) $\rightarrow$ FC(256,2) \\

\midrule
\multicolumn{2}{l}{\textbf{Collision Branch}} \\
\midrule

Path Encoder & MLP(256,512,256,24) \\
Agent Fuser & FC($256\!\cdot\!fuser\_dim$,256) \\
Agent Encoder & FC(256,256) \\
Collision Attention & Attention(256) \\
Collision Score Decoder & MLP(256,128,1) \\

\midrule
\multicolumn{2}{l}{\textbf{Traffic Light \& Stop Sign Branch}} \\
\midrule

Encoder & FC(256,256) \\
Path Encoder & FC(256,256) \\
Attention & Attention(256) \\
Score Decoder & MLP(256,128,1) \\

\bottomrule
\end{tabular}

\caption{
Detailed architecture of CLAP-AD. The fusion dimension $fuser\_dim$ is 3 for UniAD and 4 for VAD.
\textit{FC(a,b)} denotes \textit{Linear(a,b)} $\rightarrow$ LN $\rightarrow$ ReLU.
\textit{MLP(a,b,c)} denotes \textit{Linear(a,b)} $\rightarrow$ LN $\rightarrow$ ReLU $\rightarrow$ \textit{Linear(b,c)}, while
\textit{MLP(a,b,c,d)} denotes \textit{Linear(a,b)} $\rightarrow$ ReLU $\rightarrow$ \textit{Linear(b,c)} $\rightarrow$ ReLU $\rightarrow$ \textit{Linear(c,d)}.
\textit{Attention} refers to the attention module used to aggregate contextual features.
}
\label{tab:clap_architecture_detailed}
\end{table}

\paragraph{Model size.} On TFv6, the CLAP-AD head has 1.96M parameters and 0.137 GFLOPs per frame (batch size 1), about 3\% of the 61.1M-parameter model and about one fifth of TFv6s own planning decoder (9.6--10.0M parameters).

\paragraph{Hyperparameters.} The path head predicts a path in cartesian coordinates relative to the ego vehicle, and driving speed is discretized into seven classes \{0, 2.5, 5, 7, 8.5, 10, 12\}~$m/s$, chosen to cover typical urban driving behaviors in line with prior practice~\cite{jaeger2023hidden}. The collision horizon $T$ and the length of that path are not independent settings: the path has to reach at least $v_{\max}\cdot T$ for the fastest speed class to be scored over the whole horizon, which pairs $T=1\,s$ with a $12\,m$ path and $T=2\,s$ with a $24\,m$ path. We use the shorter pairing for UniAD and VAD and the longer one for TFv6. How far the horizon can usefully be extended is set by the backbone, since the supervision is only as reliable as the agent motion its features can support: on UniAD, moving to $2\,s$ lowers the Driving Score (Table 4 of the main paper), whereas TFv6 additionally consumes LiDAR and exposes a dense BEV map, which gives the collision branch the geometric evidence to remain discriminative further ahead and makes the longer pairing the better choice there. Collision scoring uses the $0.1\,m$ and $1.0\,m$ collision and safe-distance bounds given in \cref{alg:coll_algo}, and the traffic light and stop sign scores (Equation 1 of the main paper) use $a_{\mathrm{comfort}} = 2.5\,m/s^2$ and $a_{\max} = 4\,m/s^2$.

\subsection{Backbone Query Selection}
To provide CLAP-AD with the appropriate queries, we follow a selective procedure. After the detection and motion heads produce their queries along with confidence scores, we prune and filter them based on the model’s own predictions, retaining only the relevant dynamic agents and traffic elements for the corresponding CLAP-AD branches. This design leverages the perception backbone, which has already been trained standalone and performs well on its primary tasks, allowing CLAP-AD to receive reliable detection and motion queries during end-to-end training.

\subsection{Backbone Auxiliary Task}
The adaptation described here applies to UniAD alone. UniAD detects traffic lights reliably but does not distinguish their color, as that attribute is never supervised during its training. VAD and TFv6 have no such gap and are used exactly as published, without any additional supervision. The alternative would have been to alter UniAD's perception architecture, which would have stopped it from being the published baseline we compare against.

We therefore attach a small auxiliary task to UniAD's perception: predicting traffic light color and whether a detected traffic light or stop sign applies to the ego vehicle. The latter matters at complex intersections, where several traffic lights or distant and opposite-lane stop signs are visible at once. This supervision encourages the detection queries to carry the attribute the downstream scoring branches need.

In addition, we train a small classification head on UniAD's image backbone that predicts traffic light color and stop sign presence. It is trained separately on a curated subset of 62,480 Bench2Drive-base samples (out of 234,769) with a batch size of 256 and at most 2,000 update iterations. It is then frozen for both end-to-end training and inference, and its outputs are appended to the attention inputs of the traffic light and stop sign branches.

Neither component adds capability beyond what VAD and TFv6 already provide natively, and neither is part of our contribution. What they do illustrate is a side benefit of criterion-specific scoring: because each criterion is scored by its own branch, a perception deficiency surfaces as a persistently low score on a single criterion, which is how we identified this gap in the first place.

\subsection{Training Setup}
For UniAD and VAD we take the perception weights from the official Bench2Drive checkpoints and train them together with our planning head in place of the original one, fine-tuning end-to-end for one epoch. TFv6 is trained from scratch in two stages, first perception only and then end-to-end with CLAP-AD attached. All runs use AdamW with a learning rate of $2\times10^{-4}$, a batch size of 8, weight decay $0.01$ and gradient clipping at $35$, together with cosine annealing to a minimum learning-rate ratio of $1/3$ after $500$ warmup iterations.

\section{Test-Time Behavior Control on Bench2Drive}
\label{sec:controllability}
Section 4.3 of the main paper showed that CLAP-AD can follow an externally specified speed. The same mechanism also determines the model's default driving style: because behavior is set by how the controller reads the scores rather than by a fixed regressed speed, that style can be changed \emph{without retraining}. We demonstrate this on the standard 220 Bench2Drive routes.

We vary the score threshold above which a speed class is deemed safe and the rule used to pick one speed from the safe set (\cref{tab:controller_design}). A permissive controller that accepts any speed scoring above 0.3 and always selects the fastest safe class drives assertively, raising Efficiency from 176.03 to 252.84 on UniAD and from 179.01 to 251.35 on VAD. On UniAD this comes at a real cost in Driving Score (74.29 to 69.53); on VAD the cost is much smaller (75.04 to 74.54), since a lower threshold with median selection already recovers most of the achievable Driving Score for that backbone. Requiring a higher score of at least 0.5 and selecting the median safe speed instead yields cautious behavior that occasionally stalls. A balanced setting (threshold 0.3 with median selection) gives the highest Driving Score for both backbones and is the configuration reported in Table 2 of the main paper. The point is not which setting wins, but the reachable range: one trained model spans roughly 77 Efficiency points purely through how its scores are read at inference. Conventional temporal-waypoint planners expose no comparable knob, since the desired speed is implicit in the regressed trajectory and can only be altered by retraining or by post-hoc heuristics that discard the model's own safety reasoning.

\begin{table}[t]
\centering
\small
\setlength{\tabcolsep}{5pt}

\caption{\textbf{Test-time behavior control on the 220 Bench2Drive routes.} One trained model under different controller settings. The threshold sets which scores count as safe; the selection rule picks the target speed from the safe set. Varying only these inference-time parameters spans cautious to throughput-oriented driving, with no retraining.}
\label{tab:controller_design}

\begin{tabular}{lccccc}
\toprule
\textbf{Backbone}
& \textbf{Score}
& \textbf{Speed}
& \textbf{Eff.}
& \textbf{Driving}
& \textbf{Success} \\
& \textbf{Thresh.} & \textbf{Sel.} & \textbf{$\uparrow$} & \textbf{Score $\uparrow$} & \textbf{Rate(\%) $\uparrow$} \\
\midrule

\multirow{3}{*}{UniAD}
& 0.5 & median & 173.55 & 73.57 & 47.94 \\
& 0.3 & median & 176.03 & 74.29 & \textbf{49.09} \\
& 0.3 & max    & \textbf{252.84} & 69.53 & 40.45 \\
\midrule

\multirow{3}{*}{VAD}
& 0.5 & median & 172.56 & 71.21 & 42.73 \\
& 0.3 & median & 179.01 & \textbf{75.04} & 47.27 \\
& 0.3 & max    & 251.35 & 74.54 & \textbf{49.09} \\
\bottomrule

\end{tabular}
\end{table}

\section{Multi-Ability Results on Bench2Drive}
\label{sec:supp_ability}
Bench2Drive also groups its routes by the driving ability each one tests. \cref{tab:results_ability} reports the per-ability scores.

On UniAD and VAD, CLAP-AD improves every ability, and the largest changes line up with what the scoring heads evaluate. Overtaking and emergency braking gain $31.1$ and $46.7$ points on UniAD, which is where the collision head applies, and the traffic-sign ability rises from 14.21 to 55.26 on UniAD and from 19.15 to 67.89 on VAD, where the traffic-light and stop-sign heads apply. Gains that concentrate where the objective acts, rather than spreading evenly, point to the objective as their source.

On TFv6 the two groups of abilities behave differently. Traffic signs improve from 89.47 to 96.84 and give way from 40.00 to 50.00, as on the other two backbones. Overtaking and emergency braking fall, from 97.77 to 77.78 and from 91.66 to 75.00, against a baseline that already scores close to the maximum on both. This is where the 2.4 DS in Table 2 of the main paper goes, and it follows the pattern we describe in the main paper: the traffic-rule heads help on every backbone, while the collision head has room to help against a reinforcement-learning expert and very little against a rule-based one.

\begin{table}[t]
    \centering
    \caption{\textbf{Multi-ability results on Bench2Drive} (\%). On the Bench2Drive-trained backbones CLAP-AD improves every ability. On TFv6, trained on a rule-based expert, the traffic-rule abilities still improve while the collision-driven ones trade against an already strong baseline.}
    \label{tab:results_ability}
    \resizebox{\columnwidth}{!}{
    \begin{tabular}{lccccc|c}
        \toprule
        \textbf{Method} & \textbf{Merging} & \textbf{Overtaking} & \textbf{Emerg. Brake} & \textbf{Give Way} & \textbf{Traffic Sign} & \textbf{Mean} \\
        \midrule
        UniAD  & 14.10 & 17.78 & 21.67 & 10.00 & 14.21 & 15.55 \\
        CLAP-AD on UniAD  & \textbf{31.25} & \textbf{48.88} & \textbf{68.33} & \textbf{40.00} & \textbf{55.26} & \textbf{48.74} \\ \midrule
        VAD & 8.11 & 24.44 & 18.64 & 20.00 & 19.15 & 18.07 \\
        CLAP-AD on VAD & \textbf{40.00} & \textbf{42.22} & \textbf{55.00} & \textbf{60.00} & \textbf{67.89} & \textbf{53.02}\\
        \midrule
        TFv6 & \textbf{72.50} & \textbf{97.77} & \textbf{91.66} & 40.00 & 89.47 & \textbf{78.28} \\
        CLAP-AD on TFv6 & 71.25 & 77.78 & 75.00 & \textbf{50.00} & \textbf{96.84} & 74.17 \\
        \bottomrule
    \end{tabular}
    }
\end{table}

\section{Collision Score Ground-Truth Calculation}
\label{sec:supp_collision}
For each candidate speed class, we estimate the collision risk along the predicted path. Starting from the current ego position, we project the ego vehicle forward along the predicted path for a horizon $T$ according to the candidate speed. We use $T = 1\,s$ with UniAD and VAD, $T = 2\,s$ with TFv6 (\cref{sec:supp_architecture}), and the full $4\,s$ prediction window on NAVSIM. For every surrounding actor, we then compute the minimum swept bounding-box distance between the ego trajectory segment and the actor's logged motion over the same horizon, taken directly from the recorded data rather than extrapolated. The shortest distance across all valid actors is used to derive a collision score. If the minimum distance remains above a predefined safe distance of $1 m$, the motion is considered completely safe; if it falls below a distance of $0.1 m$, a collision is assumed. Otherwise, the score is linearly interpolated between these two thresholds. This procedure produces a normalized collision metric in $[0,1]$ for each speed class.

\begin{algorithm}[ht]
\caption{Collision Metric Computation for a Speed Class}
\label{alg:coll_algo}
\small
\begin{algorithmic}[1]
\Require Speed $v$, predicted path $P$, actors $\mathcal{A}$, time horizon $T$, safe distance $d_s$, collision radius $d_c$
\State Obtain ego position $p_e$
\State Compute travel distance $s \leftarrow v \cdot T$
\State Interpolate along $P$ by $s$ to obtain future ego position $p_e^{f}$
\State $d_{\min} \leftarrow d_s$
\For{each actor $a \in \mathcal{A}$}
    \If{$a$ not valid for Collision computation or logged future state unavailable}
        \State continue
    \EndIf
    \State Obtain current and logged future actor positions $(p_a, p_a^{f})$
    \State Compute swept bounding box distance $d$ between $(p_e,p_e^{f})$ and $(p_a,p_a^{f})$
    \State $d_{\min} \leftarrow \min(d_{\min}, d)$
\EndFor
\If{$d_{\min} \ge d_s$}
    \State \Return $1$ \Comment{safe}
\ElsIf{$d_{\min} \le d_c$}
    \State \Return $0$ \Comment{collision}
\Else
    \State \Return $(d_{\min}-d_c)/(d_s-d_c)$
\EndIf
\end{algorithmic}
\end{algorithm}

\section{Qualitative Results in CARLA}
\label{sec:supp_qualitative}
\begin{figure*}[t]
    \centering
    \begin{subfigure}{\textwidth}
        \centering
        \includegraphics[width=0.35\textwidth]{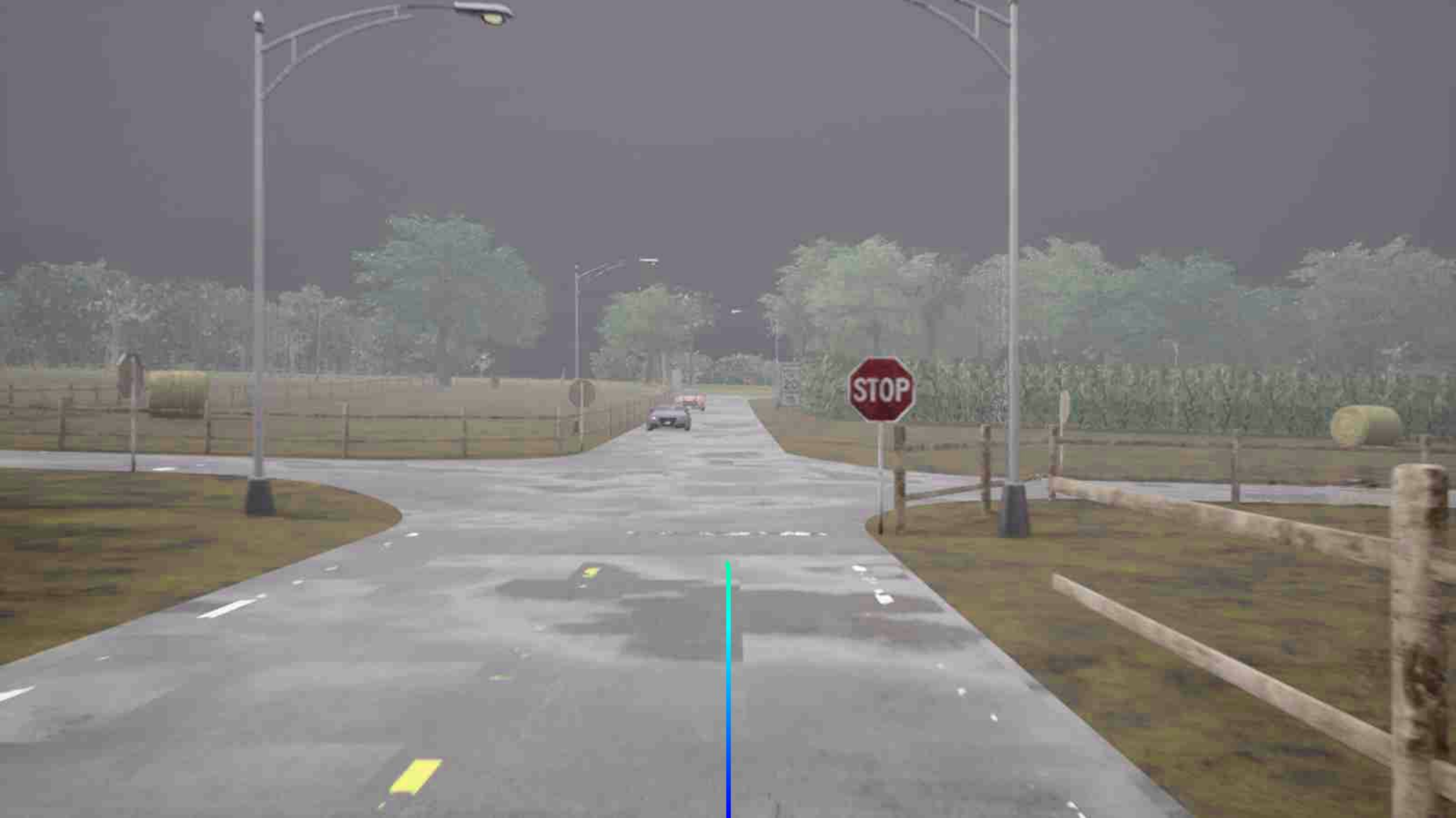}
        \includegraphics[width=0.35\textwidth]{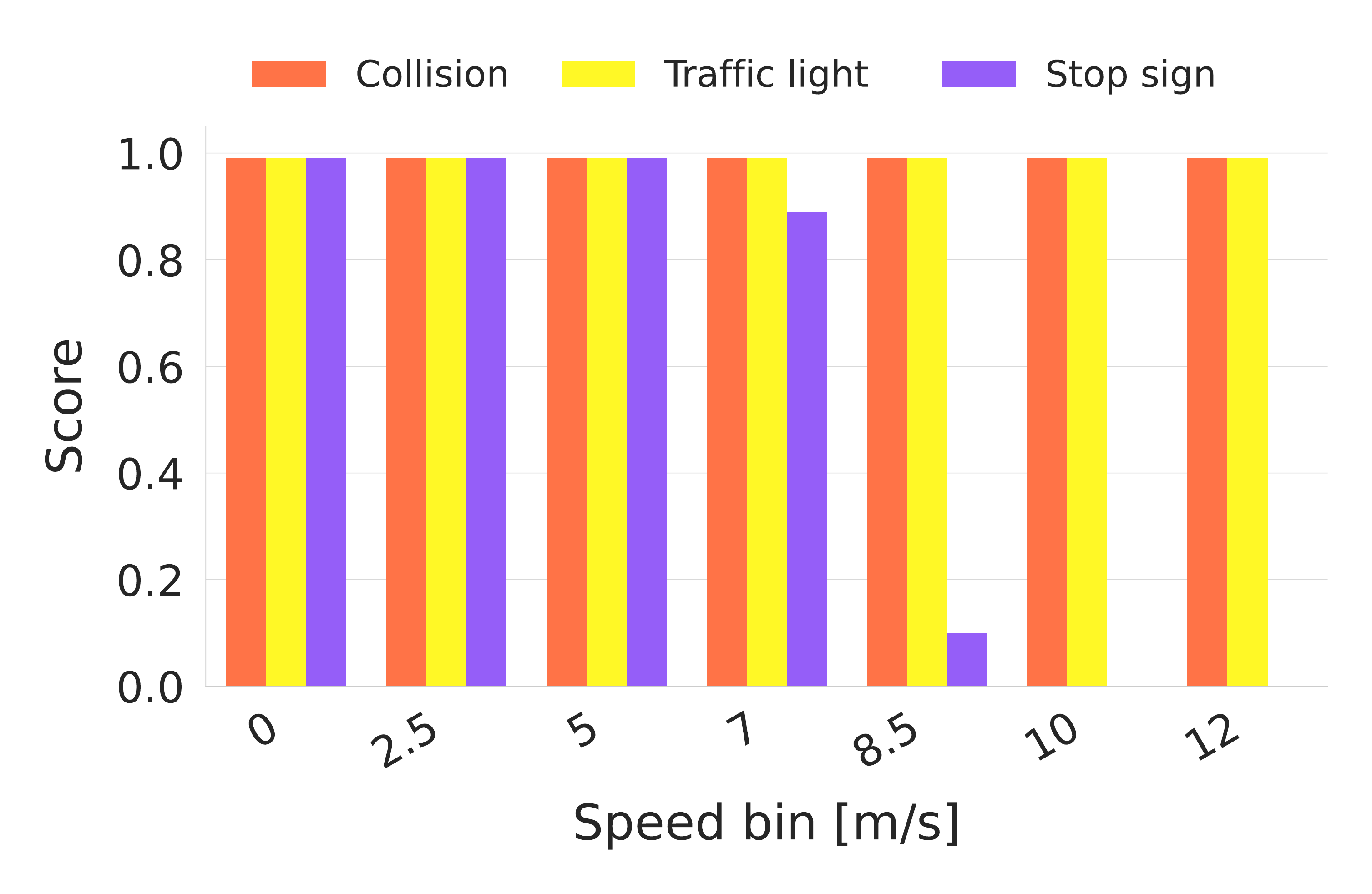}
        \caption*{\textit{Scene 1:} The ego vehicle is approaching a stop sign. The stop sign head predicts low scores for speeds over $7m/s$ as they would require excessive braking.}
    \end{subfigure}

    \begin{subfigure}{\textwidth}
        \centering
        \includegraphics[width=0.35\textwidth]{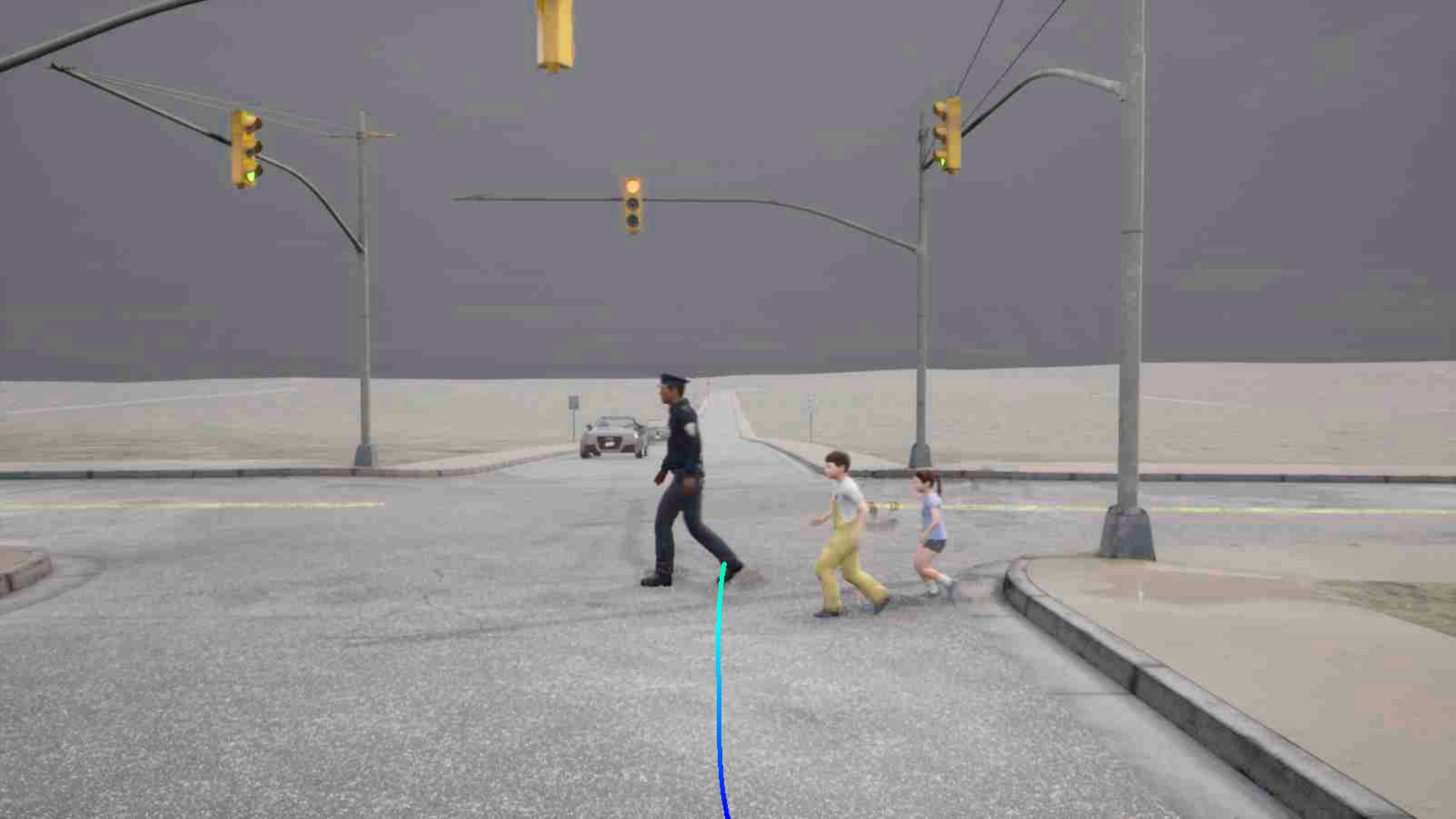}
        \includegraphics[width=0.35\textwidth]{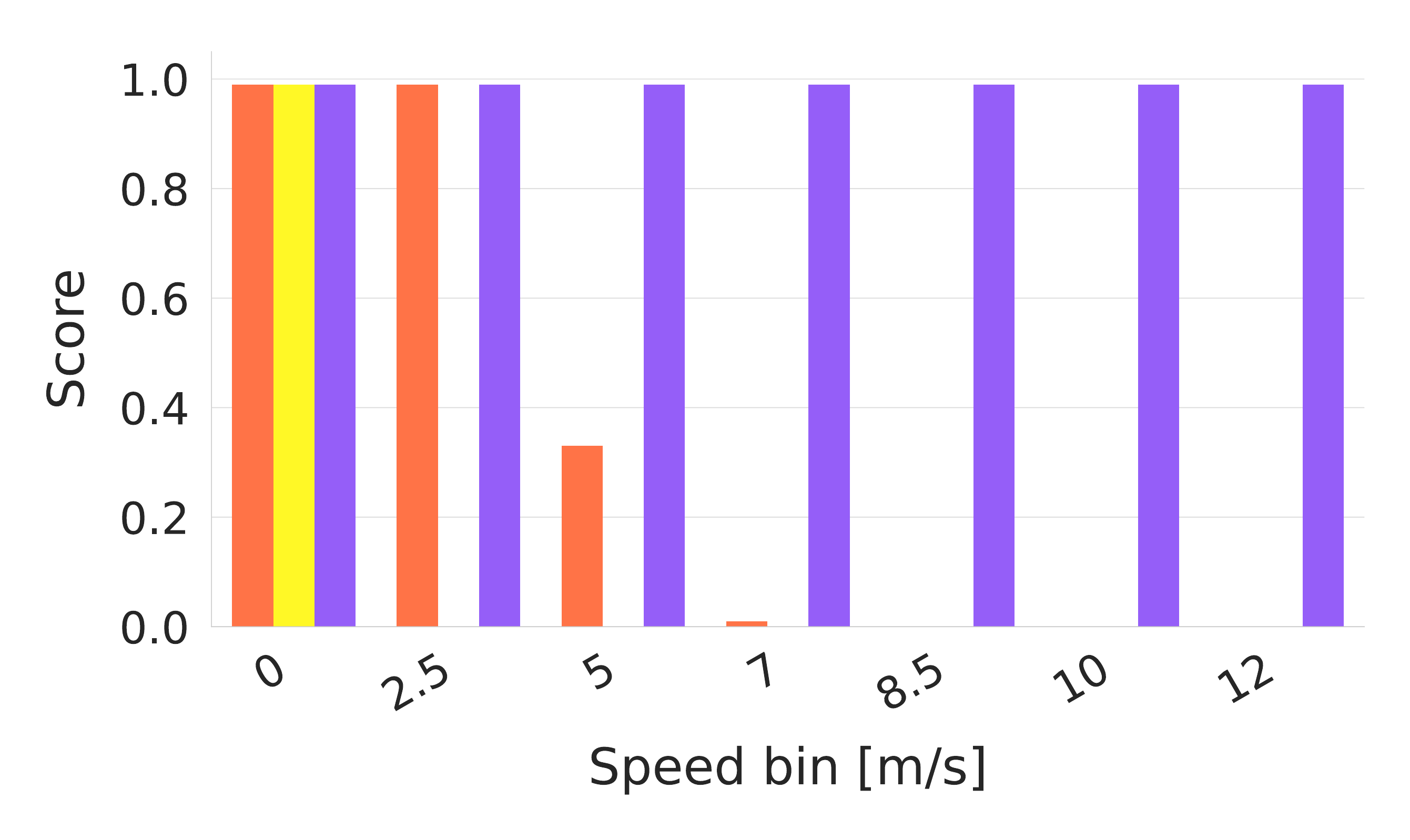}
        \caption*{\textit{Scene 2:} Pedestrians are crossing at a red traffic light. Both the collision and traffic light head predict low scores for higher speeds, as they would result in an accident and running a red light. In summary, only standing still is considered safe.}
    \end{subfigure}

    \begin{subfigure}{\textwidth}
        \centering
        \includegraphics[width=0.35\textwidth]{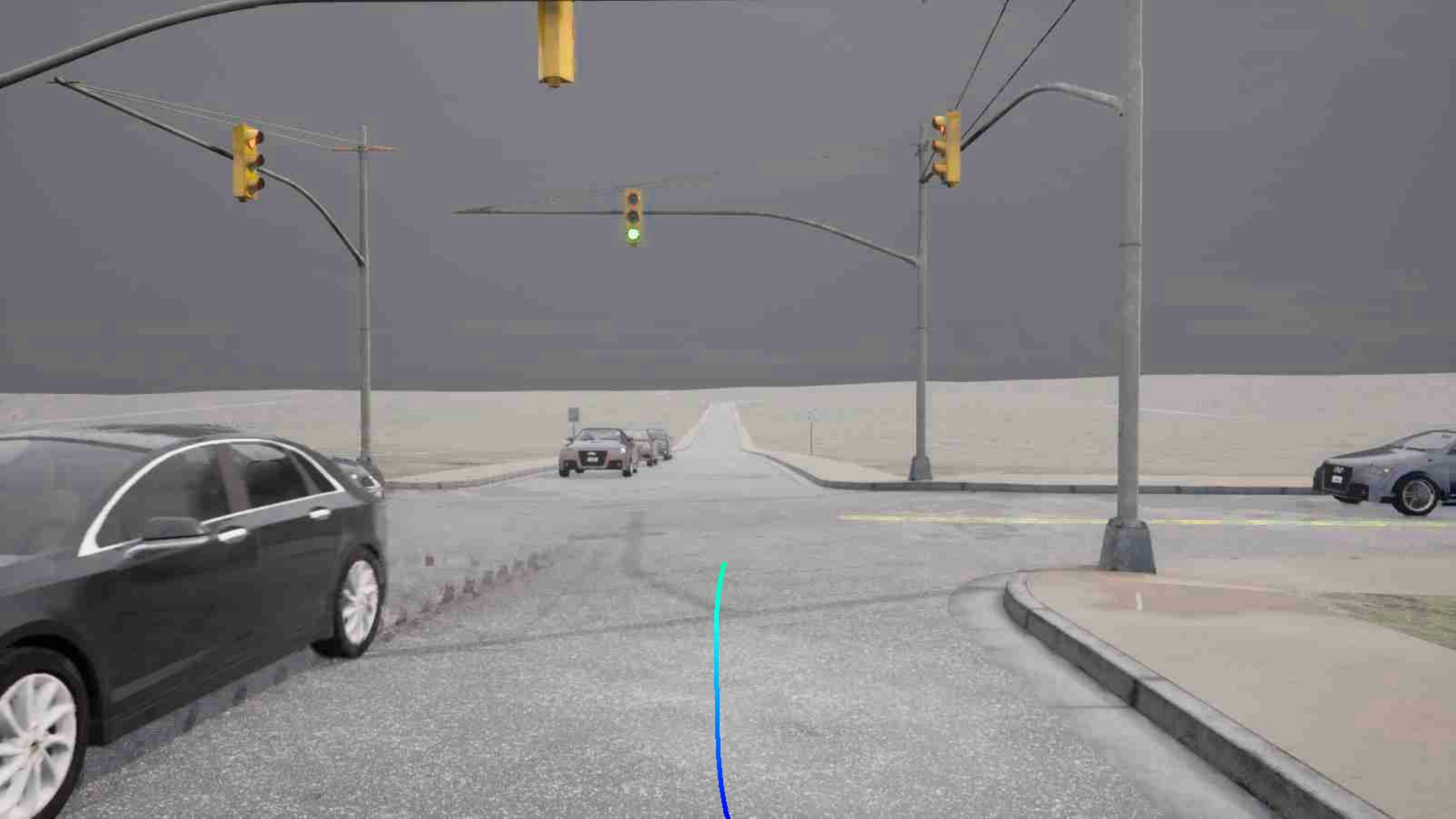}
        \includegraphics[width=0.35\textwidth]{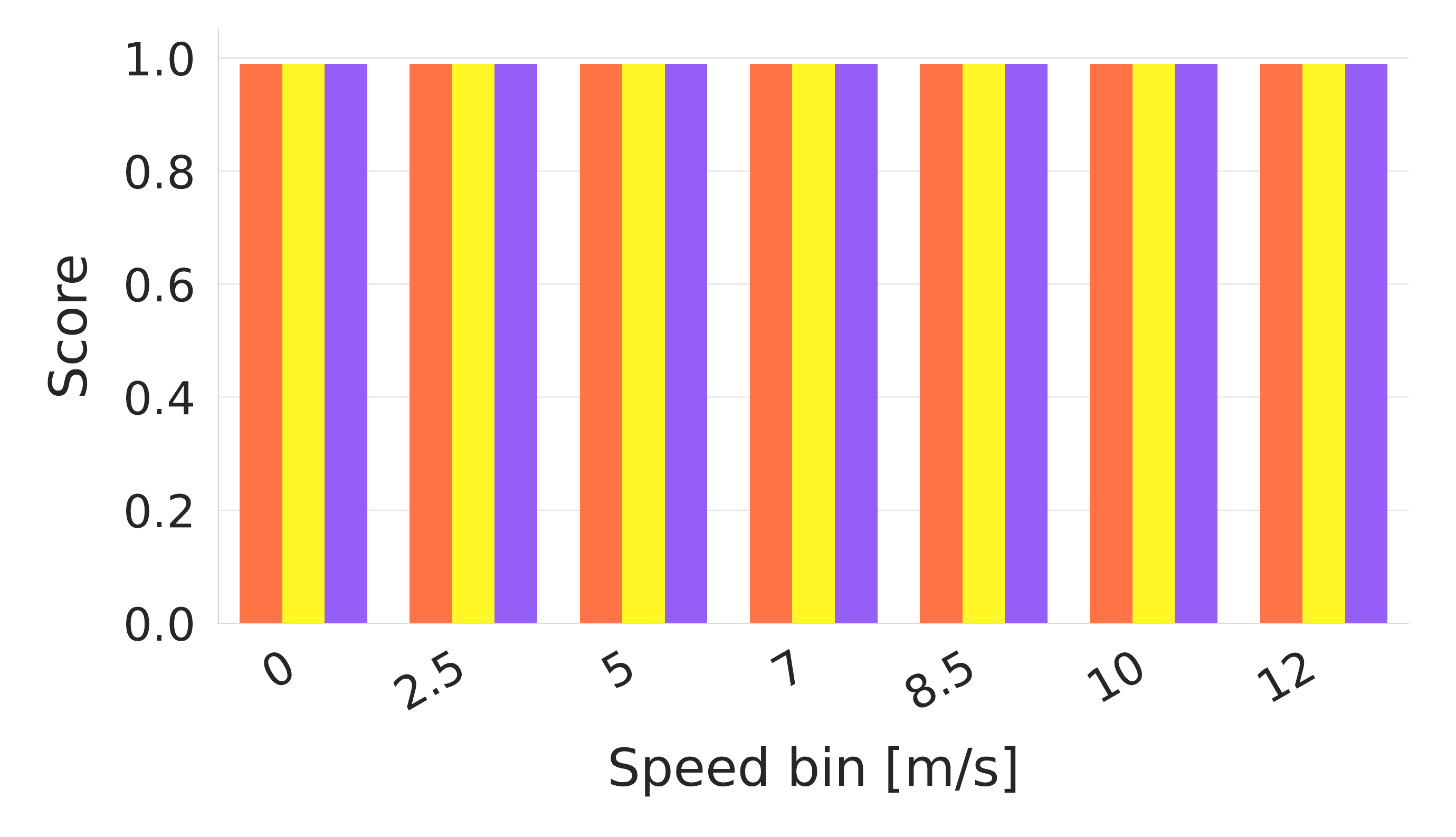}
        \caption*{\textit{Scene 3:} The traffic light is green, no stop sign is visible, and no traffic participants pose collision risk. All three heads assign high scores across the full speed range, permitting the vehicle to proceed at any reasonable speed.}
    \end{subfigure}

    \caption{Front camera images of different driving scenarios with predicted speed scores. Collision, traffic light, and stop sign branches individually predict scores for every speed class.}
    \label{fig:qualitative_rows}
\end{figure*}

\cref{fig:qualitative_rows} shows the scores CLAP-AD assigns to each speed class in three CARLA scenes, next to the front camera view at the same instant. Because every criterion is scored by its own branch (\cref{sec:supp_architecture}), the reason a speed is rejected is directly visible: the bars identify which constraint rules it out, rather than leaving that decision implicit in a regressed trajectory.

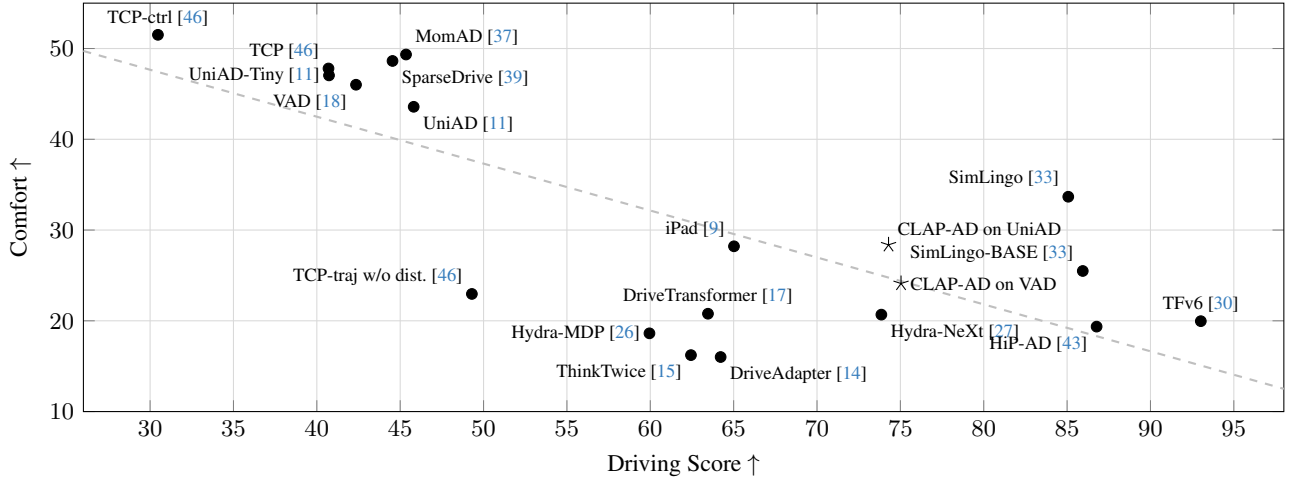
\begin{figure*}[!t]
    \centering
    \begin{tikzpicture}
        \begin{axis}[
            width=\linewidth,
            height=0.40\linewidth,
            xlabel={Driving Score $\uparrow$},
            ylabel={Comfort $\uparrow$},
            xmin=26, xmax=98,
            ymin=10, ymax=55,
            grid=both,
            minor grid style={gray!15},
            major grid style={gray!30},
            tick label style={font=\small},
            label style={font=\small},
            clip=false,
        ]

        \addplot[gray!50, dashed, thick, domain=26:98, samples=2, forget plot]
            {63.170 - 0.5170*x};

        \addplot[only marks, mark=*]
            coordinates {(40.73,47.04)}
            node[left,font=\scriptsize] {UniAD-Tiny~\cite{hu2023planning}};
    
        \addplot[only marks, mark=*]
            coordinates {(42.35,46.01)}
            node[below left,font=\scriptsize] {VAD~\cite{jiang2023vad}};
    
        \addplot[only marks, mark=*]
            coordinates {(45.81,43.58)}
            node[below right,font=\scriptsize] {UniAD~\cite{hu2023planning}};
    
        \addplot[only marks, mark=*]
            coordinates {(44.54,48.63)}
            node[below right,font=\scriptsize] {SparseDrive~\cite{sun2025sparsedrive}};
    
        \addplot[only marks, mark=*]
            coordinates {(45.35,49.34)}
            node[above right,font=\scriptsize] {MomAD~\cite{song2025don}};
    
        \addplot[only marks, mark=*]
            coordinates {(59.95,18.62)}
            node[left,font=\scriptsize] {Hydra-MDP~\cite{li2024hydra}};
    
        \addplot[only marks, mark=*]
            coordinates {(63.46,20.78)}
            node[above,font=\scriptsize] {DriveTransformer~\cite{jia2025drivetransformer}};
    
        \addplot[only marks, mark=*]
            coordinates {(65.02,28.21)}
            node[above left,font=\scriptsize] {iPad~\cite{guo2025ipad}};
    
        \addplot[only marks, mark=*]
            coordinates {(73.86,20.68)}
            node[below right,font=\scriptsize] {Hydra-NeXt~\cite{li2025hydra}};
    
        \addplot[only marks, mark=*]
            coordinates {(40.70,47.80)}
            node[above left,font=\scriptsize] {TCP~\cite{wu2022trajectory}};
    
        \addplot[only marks, mark=*]
            coordinates {(30.47,51.51)}
            node[above,font=\scriptsize] {TCP-ctrl~\cite{wu2022trajectory}};

        \addplot[only marks, mark=*]
            coordinates {(49.30,22.96)}
            node[above left,font=\scriptsize] {TCP-traj w/o dist.~\cite{wu2022trajectory}};
    
        \addplot[only marks, mark=*]
            coordinates {(62.44,16.22)}
            node[below left,font=\scriptsize] {ThinkTwice~\cite{jia2023think}};
    
        \addplot[only marks, mark=*]
            coordinates {(64.22,16.01)}
            node[below right,font=\scriptsize] {DriveAdapter~\cite{jia2023driveadapter}};
    
        \addplot[only marks, mark=star, mark size=3pt]
            coordinates {(74.29,28.37)}
            node[above right,font=\scriptsize] {CLAP-AD on UniAD};
    
        \addplot[only marks, mark=star, mark size=3pt]
            coordinates {(75.04,24.11)}
            node[right,font=\scriptsize] {CLAP-AD on VAD};
    
        \addplot[only marks, mark=*]
            coordinates {(85.94,25.49)}
            node[above left,font=\scriptsize] {SimLingo-BASE~\cite{renz2025simlingo}};
    
        \addplot[only marks, mark=*]
            coordinates {(85.07,33.67)}
            node[above left,font=\scriptsize] {SimLingo~\cite{renz2025simlingo}};
    
        \addplot[only marks, mark=*]
            coordinates {(86.77,19.36)}
            node[below left,font=\scriptsize] {HiP-AD~\cite{tang2025hip}};
    
        \addplot[only marks, mark=*]
            coordinates {(93.02,19.96)}
            node[above,font=\scriptsize] {TFv6~\cite{nguyen2026lead}}; %
    
        \end{axis}
    \end{tikzpicture}
    \caption{\textbf{Driving Score versus Comfort on the 220 Bench2Drive routes.} Comfort falls as Driving Score rises; the dashed line is a least-squares fit over the methods shown. This is not evidence that competent driving is uncomfortable: every quantity the metric monitors is an amplitude of motion, so policies that stall or creep satisfy all of them trivially.}
    \label{fig:driving_score_comfort}
\end{figure*}

\section{NAVSIM Evaluation and Trajectory Generation}
\label{sec:navsim_supp}

\paragraph{EPDMS and the two evaluation stages.}
NAVSIM~v2~\cite{cao2025pseudo} is a pseudo-simulation benchmark: it evaluates on real logged data, as in open-loop evaluation, but augments each scene with synthetic observations rendered ahead of time by 3D Gaussian Splatting from perturbed ego states. This probes recovery from error without running an interactive simulator. Evaluation proceeds in two stages. In Stage~1 the agent plans a four-second trajectory from the real observation. In Stage~2 it re-plans from roughly twelve synthetic observations per scene, taken from positions, headings and speeds around the Stage~1 endpoint, and those scores are combined by a Gaussian-weighted average favoring viewpoints close to the trajectory the agent actually predicted. The reported EPDMS is the product of the two stage scores, so a method must both plan well from the logged state and remain sensible when started from a state it did not choose.

Each stage score is assembled from nine sub-metrics. No-at-fault collision (NC), drivable-area compliance (DAC), driving-direction compliance (DDC) and traffic-light compliance (TL) enter multiplicatively, so failing any one of them zeroes the score for that scene. Ego progress (EP), time-to-collision (TTC), lane keeping (LK), history comfort (HC) and extended comfort (EC) then enter a weighted average, with weights $5$, $5$, $2$, $2$ and $2$ respectively.

\paragraph{Trajectory generation from CLAP-AD outputs.}
CLAP-AD predicts a spatial path $\{p_i\}_{i=1}^{T}$ in the ego frame together with per-bin collision-safety logits over a discrete set of speed bins $\{v_s\}$. We convert these into a kinematically feasible $4\,\mathrm{s}$ trajectory as follows. First, a soft target speed $v^\ast$ is obtained by weighting each bin by a sigmoid confidence score centered on a safety threshold rather than a hard cutoff, $v^\ast = \sum_s w_s v_s / \sum_s w_s$ with $w_s = \sigma((\mathrm{score}_s - \tau)/T)$. This keeps the target speed continuous under small changes in the initial state, which is what the Stage~2 viewpoints vary and what the extended comfort sub-metric is sensitive to. Second, we roll out a speed profile from the current speed $v_0$ toward $v^\ast$ under bounded acceleration and braking limits of $0.7$ and $1.4\,\mathrm{m/s^2}$, matched to the traffic model's own kinematic limits; if the resulting profile would overshoot the distance implied by holding $v^\ast$ for the full horizon, we binary-search for the largest feasible ramp target that respects this distance budget. Finally, the cumulative arc length of the speed profile is computed by trapezoidal integration and used to resample the predicted path at the corresponding arc-length positions, yielding the final $(x,y)$ waypoints, with heading at each waypoint taken from the local path tangent.

\section{Bench2Drive Evaluation Metrics}
\label{sec:supp_metrics}
We evaluate our models using the closed-loop metrics proposed by Bench2Drive.
\textbf{Success Rate (SR)} measures the proportion of routes completed without traffic violations within the allotted time. A route is considered successful if the ego vehicle reaches its destination while adhering to all traffic rules.
\textbf{Driving Score (DS)} follows the CARLA official metric, combining route completion with penalties for infractions. DS averages route completion percentages and applies severity-weighted penalties for violations, normalized over routes of the same type or group.
\textbf{Efficiency} evaluates how well the vehicle maintains speed relative to nearby traffic. Bench2Drive extends CARLA’s original metric by using 20 evenly spaced checkpoints along each route (every 5\% of the total route length) to compute the ego vehicle’s speed as a percentage of surrounding traffic, averaging over all checkpoints. Unlike the CARLA Leaderboard, which folds the speed check into the Driving Score as a penalty, Bench2Drive excludes it, so Efficiency and Driving Score are reported independently.
Bench2Drive additionally reports \textbf{Comfort}, which we define and discuss separately in \cref{sec:comfort}.

\section{On the Comfort Metric}
\label{sec:comfort}

\textbf{Comfort} follows nuPlan's smoothness protocol. At every frame, Bench2Drive checks the ego's minimum and maximum longitudinal acceleration, maximum absolute lateral acceleration, yaw rate, yaw acceleration, longitudinal jerk and total jerk magnitude against bounds fitted to human expert trajectories, and marks the frame comfortable only if all of them fall inside their bounds. The reported score is the fraction of comfortable frames.

We omit this metric from the comparisons in the main paper. This is not because it disadvantages us: of the seven methods above 70 Driving Score in \cref{fig:driving_score_comfort}, our two variants rank second and fourth on Comfort. We omit it because, as defined, it does not measure what its name suggests. As analyzed in~\cite{aghajanzadeh2026comfortablystuckevaluationpitfalls}, across eight models the metric correlates with the time a vehicle spends stationary at $r=0.97$, owing to implementation errors and a time-based segmentation that rewards standing still; a distance-based correction reduces this correlation to $r=0.06$.

Every quantity the metric monitors is an amplitude of motion, so a policy that barely moves satisfies all of them trivially. A vehicle that stalls at a junction, creeps behind an obstacle, or ends its route early produces near-zero accelerations, yaw rates and jerks, and is scored as maximally comfortable. \cref{fig:driving_score_comfort} shows the consequence across twenty published methods, which fall on a clear downward trend ($r = -0.73$): the highest Comfort in the entire comparison, 51.51, belongs to TCP-ctrl, one of the weakest drivers in it at 30.47 Driving Score, while the strongest drivers cluster near 20. The trend is therefore not evidence that competent driving is uncomfortable. It is evidence that the metric rewards inaction, which is precisely the failure mode Bench2Drive was built to expose.

Two further properties compound this. The thresholds are transplanted from nuPlan, where they were fitted to ordinary human driving, whereas Bench2Drive consists of edge-case scenarios in which the correct response is frequently a hard brake or a sharp avoidance; the maneuver that earns Driving Score is by construction outside the bounds. And because each frame is scored by a hard conjunction of booleans, a marginal overshoot on one variable counts exactly as much as a violent one, so the metric cannot separate a firm but controlled brake from an erratic one.

We therefore regard Comfort on Bench2Drive as uninformative for ranking planners and report Driving Score, Success Rate and Efficiency instead.

\section{Limitations}
\label{sec:limitations}
The scoring in CLAP-AD covers three criteria, collision, traffic lights and stop signs, and it does not consider road speed limits, for example on highways. The candidate speed set is chosen to cover typical urban driving (\cref{sec:supp_architecture}), so higher-speed settings such as highways would require both an extended speed range and a criterion that scores speeds against the applicable limit. A second limitation is that for some complicated maneuvers it may be difficult to formulate a scoring mechanism at all, since the target for each speed class has to be derived from an explicit, checkable rule. Scoring is therefore best used for hard, strict rules, such as keeping a safe distance or stopping at a red light, where it provides transparent and interpretable safety checks and lets failures be attributed to individual criteria. For more complicated maneuvers, expert behavior can be taken into account additionally, while the scores continue to take care of these safety checks, so that the two sources of supervision complement each other.

\end{document}